\documentclass[10pt,twocolumn,letterpaper]{article}

\usepackage[pagenumbers]{cvpr} 

\usepackage{graphicx}
\usepackage{amsmath}
\usepackage{amssymb}
\usepackage{booktabs}  

\usepackage[pagebackref,breaklinks,colorlinks]{hyperref}

\usepackage[capitalize]{cleveref}
\crefname{section}{Sec.}{Secs.}
\Crefname{section}{Section}{Sections}
\Crefname{table}{Table}{Tables}
\crefname{table}{Tab.}{Tabs.}

\def\cvprPaperID{2743} 
\def\confName{CVPR}
\def\confYear{2022}

\usepackage{enumitem} 
\usepackage{overpic} 
\usepackage{color}
\usepackage{times}
\usepackage{epsfig}
\usepackage[utf8]{inputenc} 
\usepackage{amsfonts}       
\usepackage{nicefrac}       
\usepackage{algpseudocode}
\usepackage{multirow}
\usepackage{mathtools}
\usepackage{changepage}
\usepackage{color, colortbl}
\usepackage{arydshln}  
\usepackage{enumitem}  
\usepackage{algorithm}
\usepackage{url}
\usepackage{hyperref}
\usepackage{bbm}
\usepackage{array}
\usepackage{xcolor}
\usepackage[normalem]{ulem}

\usepackage{microtype} 
\usepackage{placeins} 
\definecolor{turquoise}{cmyk}{0.65,0,0.1,0.3}
\definecolor{purple}{rgb}{0.65,0,0.65}
\definecolor{dark_green}{rgb}{0, 0.5, 0}
\definecolor{orange}{rgb}{0.8, 0.6, 0.2}
\definecolor{red}{rgb}{0.8, 0.2, 0.2}
\definecolor{darkred}{rgb}{0.6, 0.1, 0.05}
\definecolor{blueish}{rgb}{0.0, 0.3, .6}
\definecolor{light_gray}{rgb}{0.7, 0.7, .7}
\definecolor{pink}{rgb}{1, 0, 1}
\definecolor{greyblue}{rgb}{0.25, 0.25, 1}

\hypersetup{urlcolor=blue}
\makeatletter
\DeclareUrlCommand\ULurl@@{%
  \def\UrlLeft{\uline\bgroup}%
  \def\UrlRight{\egroup}}
\def\ULurl@#1{\hyper@linkurl{\ULurl@@{#1}}{#1}}
\DeclareRobustCommand*\ULurl{\hyper@normalise\ULurl@}
\makeatother

\DeclarePairedDelimiter{\norm}{\lVert}{\rVert}
\DeclareMathOperator*{\argmax}{argmax} 
\definecolor{Gray}{gray}{0.95}

\usepackage{blindtext}

\renewcommand{\paragraph}[1]{\vspace{1em}\noindent\textbf{#1}.}
\begin{document}
\title{Simple Post-Training Robustness \\ Using Test Time Augmentations and Random Forest}

\author{Gilad Cohen and Raja Giryes\\
Tel Aviv University\\
Tel Aviv, 69978\\
{\tt\small \{giladco1@post,raja@tauex\}.tau.ac.il}
}
\maketitle
\begin{abstract}
Although Deep Neural Networks (DNNs) achieve excellent performance on many real-world tasks, they are highly vulnerable to adversarial attacks. A leading defense against such attacks is adversarial training, a technique in which a DNN is trained to be robust to adversarial attacks by introducing adversarial noise to its input. This procedure is effective but must be done during the training phase. In this work, we propose Augmented Random Forest (ARF), a simple and easy-to-use strategy for robustifying an existing pretrained DNN without modifying its weights. For every image, we generate randomized test time augmentations by applying diverse color, blur, noise, and geometric transforms. Then we use the DNN's logits output to train a simple random forest to predict the real class label. Our method achieves state-of-the-art adversarial robustness on a diversity of white and black box attacks with minimal compromise on the natural images' classification. We test ARF also against numerous adaptive white-box attacks and it shows excellent results when combined with adversarial training. Code is available at \ULurl{https://github.com/giladcohen/ARF}.
\end{abstract}
\section{Introduction}
\label{sec:intro}
Deep neural networks (DNNs) achieve cutting edge performance in many problems and tasks. Yet, it has been shown that small perturbations of the network, which in many cases are indistinguishable to a human observer, may alter completely the network output \cite{FGSM,Intriguing}. This phenomenon poses a great risk when using neural networks in sensitive applications and therefore requires a lot of attention.  

Many defense techniques were developed to improve DNN's robustness to adversarial attacks. Yet, repeatedly, after the proposal of a new successful defense, a new attack was proposed that found new breeches in the DNNs \cite{Carlini2017Bypassing,Tramer2020Adaptive,Athalye2018ObfuscatedGG}.

An example of a very common and successful strategy for improving DNN robustness is adversarial training \cite{FGSM,Towards_DeepLearning_Resistance}. In this approach, adversarial examples are added in the network training process along with the regular examples. It is shown to reduce significantly the network vulnerability to attacks.
\begin{figure}
\centering
\includegraphics[width=\linewidth]{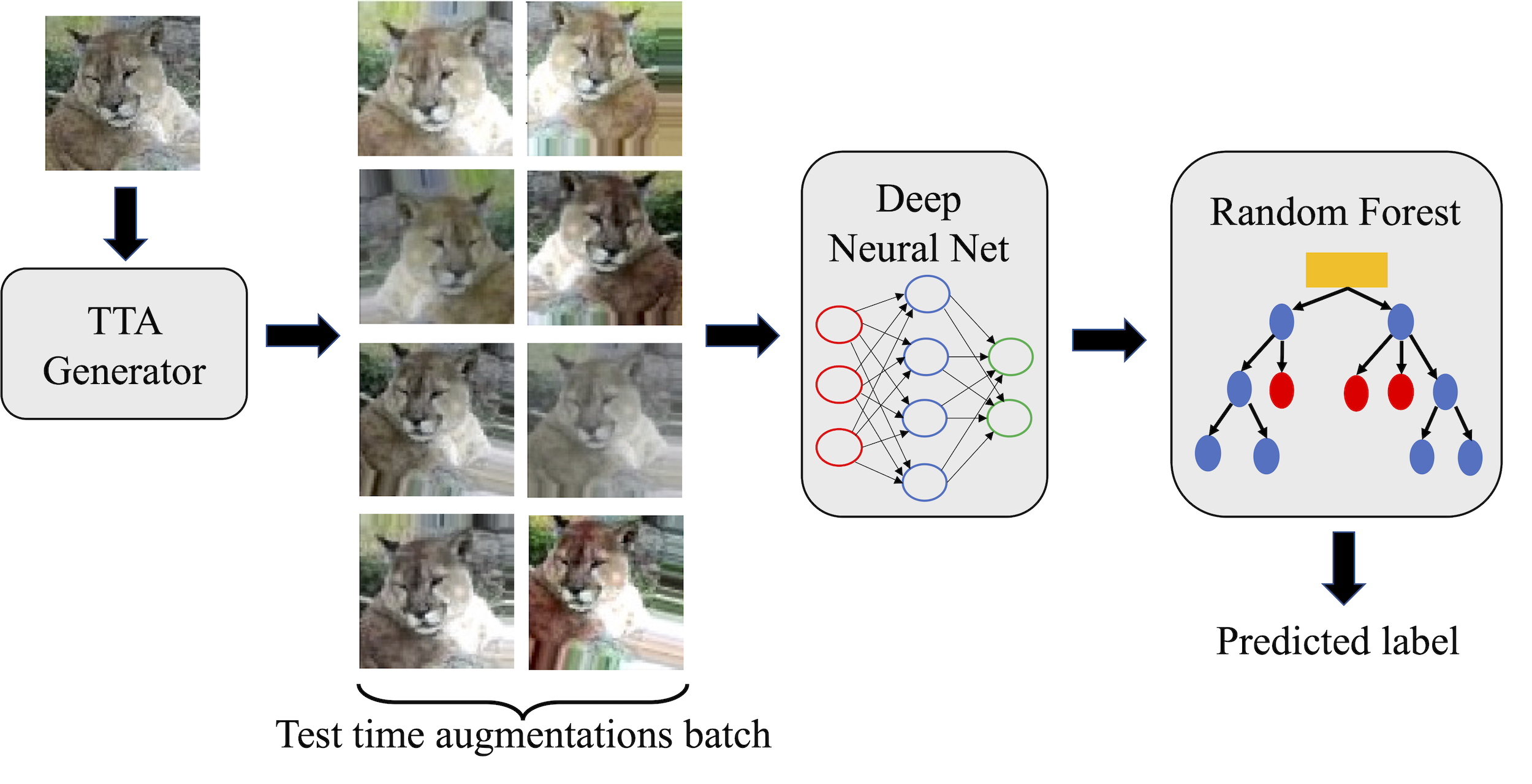}
\caption{\textbf{ARF flow chart}. Test time augmentations are generated and fed into a pretrained DNN. Its logits are then passed to a random forest classifier to predict the class label.}
\label{fig:arf_flow}
\vspace{-0.1in}
\end{figure}

A major disadvantage of this approach and most of the other existing defense strategies is that they require retraining the network. This puts an additional computational time, which might be significant in some cases, as one needs to update the network frequently to resist novel attacks.

Even in techniques that just fine tune DNNs, there is a need to have an access to all the training data. The same holds for the current leading detection methods that aim at just spotting attacks and alerting about them (without changing the DNN) \cite{LID,Mahalanobis_adv_detection,Cohen_2020_CVPR}.
Besides storage issues, having access to all data is a problem when a user wants to improve its network robustness to new attacks that were not present during the development of the DNN but does not have access to that data due to privacy or proprietary issues.

{\bf Contribution.} To mitigate these issues, we propose a novel approach for improving the robustness against adversarial attacks, named \textbf{A}ugmented \textbf{R}andom \textbf{F}orest (ARF), which only requires storage of logits vectors and not the images themselves. Also, it is very simple to use, does not require retraining, and can be employed with any machine learning classifier that produces logits, including an adversarially trained DNN, to improve its robustness. In addition to our proposed defense, we also introduce a novel white-box attack, A-PGD, that combines the PGD attack \cite{Towards_DeepLearning_Resistance} with image augmentations. We show this attack is superior than current state-of-the-art (SOTA) attacks on DNN ensembles.

In our approach, for each input image we generate $N$ Test Time Augmentations (TTAs) as shown in Figure~\ref{fig:arf_flow}. We feed this batch of image transformations to the DNN and collect its logits output.
In the training phase (done only once), we fit a simple random forest classifier using logits obtained from both natural and adversarial TTAs. The random forest learns to be robust against adversarial images by training on the entire set of TTAs' logits distribution. In the inference phase, we generate $N$ TTAs for a single image, obtain the DNN logits for it, and pass the logits to the random forest classifier. The inference executes a single forward pass on all the $N$ TTAs at once, and thus is very fast.

We emphasize that the image transformations done in our pre-processing were used in previous works for defense \cite{Nesti2021DetectingAE,guo2018countering,Luo2015FoveationbasedMA,Wang2016LearningAD}. These TTAs improve classifiers' robustness due to obfuscated gradients, but this was later shown to be fake robustness which can be circumvented using adaptive attacks in white-box settings \cite{Athalye2018ObfuscatedGG} (see Section \ref{sec:related}). Yet, we utilize TTAs to enrich the input augmentations for the random forest; we find that they improve accuracy on natural (non-adversarial) images and also enhance the robustness.


We compare the adversarial robustness of ARF to SOTA baselines on a diverse set of attack strategies and threat models, for CIFAR-10, CIFAR-100, SVHN, and Tiny-ImageNet.
ARF always succeeds to enhance the DNN's robustness by many folds, without the need to retrain the network (the random forest training is negligible compared to a neural network training) or store the training data. The best results are obtained when applying ARF with an adversarially robust network. We show that this combination achieves SOTA defense and is robust to adaptive white-box attacks.

\section{Related works}
\label{sec:related}
Various attacks and defense techniques have been proposed for DNNs. Defense techniques may be divided into strategies that aim at increasing the network robustness and approaches that try only detecting the adversarial attacks; In this work we focus on the former ones and describe some of them. For a more a comprehensive survey of the existing strategies one may refer to \cite{Rouani19Safe,Yuan19Adversarial,Miller20Adversarial}. We start by describing some of the existing attack strategies. 

{\bf Adversarial attacks.} The core strategy in adversarial attacks is to look for the smallest perturbation of an input that causes the network to change its prediction. The main difference between different existing attacks is the metric used to define the size of the change and the search strategy that is used for finding the perturbation. In addition, attacks may be targeted, i.e., aiming to change the output to a specific given class, or untargeted that just try to flip the network prediction. Another difference is the threat model of the attacks. Black-box adversaries can merely access the network outputs, while white-box adversaries have full access to the network architecture and parameters, algorithm used for training and classification, and training data \cite{Chakraborty2018AdversarialAA}.

We turn to describe some of the existing attacks. 
The fast gradient sign method (FGSM) changes the input in the direction of the gradient of the cross-entropy loss \cite{FGSM}. It is a fast single-step attack that is very easy to deploy.
The Jacobian-based saliency map attack (JSMA) aims to find only a selected few input pixels (i.e., it uses the $L_0$ metric), which induce the largest loss increase, and modify only them \cite{JSMA} . It is stronger but iterative and slow.

Deepfool is a non-targeted attack that searches for the closest decision boundary of the network for the given input example \cite{DeepFool}. 
Carlini and Wagner\cite{CarliniWagner2017Towards} proposed a novel targeted attack (known as CW due to the authors name), which was able to overcome the distillation defense method that was very successful till then \cite{Distillation}.
Their approach was further improved in \cite{Carlini2017Bypassing}, where they formulated an optimization framework to construct loss functions for attacks that are defense specific, i.e., adapted to specific defenses at hand.
Madry et. al showed that Projected Gradient Decent (PGD) is an optimal first order adversary, and applying it on a DNN during an adversarial training achieves optimal robustness against any first order attack \cite{Towards_DeepLearning_Resistance}.

Brendel et. al \cite{Boundary2018} introduced the Boundary attack, a decision based attack used in black-box settings. Their method only requires the final model prediction and can be employed where the output logits do not exist or inaccessible.
A more efficient black-box attack was introduced by Andriushchenko et. al \cite{SquareAttack}. They used much fewer queries than the Boundary attack, and it was shown to even outperform several gradient-based white-box attacks. These two attacks do not rely on gradient information and can be applied on any machine learning classifier.

In order to evaluate the performance of a novel defense approach, it is not sufficient to check robustness on the above attacks but rather design adaptive attacks to the developed defense \cite{OnEvaluatingAdvRobustness2019}. In our work, we evaluate our proposed defense against several tailored adaptive attacks. 

\textbf{Adversarial robustness.}
Many techniques where proposed to improve the adversarial robustness of DNNs.

Some techniques add a regularization during the network training such as penalizing the network input gradients to improve robustness \cite{InputGradients}; penalizing the network Jacobian showing that it increases the decision boundary and thus improve robustness \cite{Jakubovitz2018ImprovingDR}; scaling the gradients in a batch based on their magnitude \cite{BANG}; penalizing the network output so it has a smaller Lipschitz constant \cite{Cross_Lipschitz_regularization}; requiring similarity between logits of pairs of input examples \cite{kannan2018adversarial}; requiring the linear and convoluional layers in the network to be approximately Parseval tight frames \cite{Parseval}; or using the mixup regularization \cite{zhang2017mixup,zhang2020does}. 

Other approaches rely on gradient masking \cite{samangouei2018defensegan,Dhillon2018stochastic,buckman2018thermometer,guo2018countering}, i.e., they make it harder for attacks (especially black-box) to be able to find the gradient direction for producing the adversarial examples.
However, masking the model gradient's cannot guarantee robustness in white-box setting against adaptive attacks, as shown by Athalye et. al \cite{Athalye2018ObfuscatedGG}. They proposed the BPDA attack which estimates masked gradients in the classifier, and replaces them with approximated gradients in the backward pass by replacing any non-differential layer.

Another strategy to improve robustness is adding noise to the data or perturbations to the network features during training \cite{Jeddi_2020_CVPR,xie2019feature,Cohen19Certified}.
A different approach performs a $k$-NN search, perhaps using external datasets or the web, to make a decision on the input \cite{Dubey2019DefenseAA,Sitawarin2019DefendingAA}. 

Papernot et. al proposed to use knowledge distillation to improve adversarial robustness \cite{Distillation}. It was demonstrated that when one wishes to distill a robust network weights to another model (even with different architecture), the information of the gradients can improve the transfer \cite{Chan_2020_CVPR,DBLP:journals/corr/PapernotMG16}.

A leading method is adversarial (re)training with its many variants \cite{FGSM,Towards_DeepLearning_Resistance,Kurakin2017AdversarialML,tramer2018ensemble,Shaham2018RobustOptimization,Virtual_Adversarial_Training,wong2019fast}. It trains the network using adversarial examples in addition to the regular data and thus improves its robustness. 
It was suggested to add unlabeled data in the adversarial training to improve performance on the clean data \cite{carmon2019unlabeled,zhai2019adversarially}, which is deteriorated many times due to the adversarial training.

One of the disadvantages of many of the adversarial training methods is that they are computationally demanding. To alleviate that, the ``free adversarial training'' approach that is fast and leads to robust networks was suggested \cite{shafahi2019adversarial}.

Miyato et. al proposed the Virtual Adversarial Training (VAT) \cite{Virtual_Adversarial_Training}. They used a regularization term to smooth the output logits distribution of the model within a small environment surrounding the input image. Note that it is not adversarial training \textit{per-se} as it does not introduce adversarial images and its regularization does not require labeling, thus making it suitable also for semi-supervised learning.

Zhang et. al proposed TRADES, adding a regularizaion term to the cross-entropy loss in the training phase to improve robustness inside the $\ell_p$ ball $\mathbb{B}_p(x, \epsilon) = \{x': ||x - x'||_p \leq \epsilon\}$ \cite{TRADES}. By adjusting this term one can control the trade-off between the accuracies on the normal and adversarial samples \cite{Stutz_2019_CVPR}.
It was demonstrated that using robust training in the self-supervised learning regime leads to network robustness also after it is being fine-tuned for the down-stream tasks \cite{Chen_2020_CVPR,Jiang2020Robust}.

Notice that all of the above approaches require changing the DNN training and cannot be used for an already trained network. Indeed, one may use feature squeezing techniques that smooth or quantize the DNN input \cite{Xu2018FeatureSD}, but these methods are weaker than TRADES or VAT.

\textbf{Test-time augmentation (TTA).}
Some works used transformations on the input image to yield a robust classifier \cite{Luo2015FoveationbasedMA,Lu2017NONT,Dziugaite2016ASO,Graese2016AssessingTO,Wang2016LearningAD,xie2018mitigating}. However, Athalye et. al found that these approaches are susceptible to the EoT attack in white-box settings \cite{Athalye2018SynthesizingRA}, where the distribution of the transformation is considered in the attack loss. In a later work,  the BPDA attack was demonstrated to circumvent non-differential transformations as well \cite{Athalye2018ObfuscatedGG}.

Using image augmentations in test time was proved useful for many tasks out of the scope of adversarial robustness. \cite{Sun2020TestTimeTW} employed TTAs to improve generalization of out-of-distribution images in test time, by fine tuning the model's parameters before making a prediction.
SimCLR \cite{SimCLR} showed (among many other self-supervised approaches) that extensive data augmentation as implemented by us in Section~\ref{sec:test_time_augmentations} are very useful for self-supervised learning.


In \cite{guo2018countering,bahat2019natural} the KL divergence between a pair of augmentations was computed to detect adversarial attacks. The work in \cite{Roth2019TheOA} also proposed to utilize TTAs to detect adversarial images. The TTAs were used to aggregate statistics on the input image and detect anomalies associated with adversarial attacks. They showed that in some cases the correct label can also be predicted. Their approach requires an extensive statistical analysis on the dataset and tuning parameters and thresholds. Thus, it is not simple and easy to use with any arbitrary pretrained classifier.

The closest work to ours introduced a random forest classifier after a DNN for improving robustness to adversarial attacks \cite{Ding19Defending}. Unlike our defense, their methodology includes a tedious analysis on the DNN layers. They searched for the "best" layer to start growing the random forest using a manual observation on the relative $L_2$ distance between original and adversarial samples, whereas our method simply attaches the output of any machine learning classifier to a vanilla random forest. This work showed robustness on simple MNIST and CIFAR-10 datasets, whereas our work also exhibits robustness on Tiny Imagenet, a much more complex dataset.
In addition, their method does not show robustness to white-box attacks although their masked gradients can be estimated using BPDA. Our Defense is extensively tested on a variety of threat models, including a white-box setting.
\section{Method}
\label{sec:method}

\begin{figure}[t!]
\centering
\includegraphics[width=\linewidth]{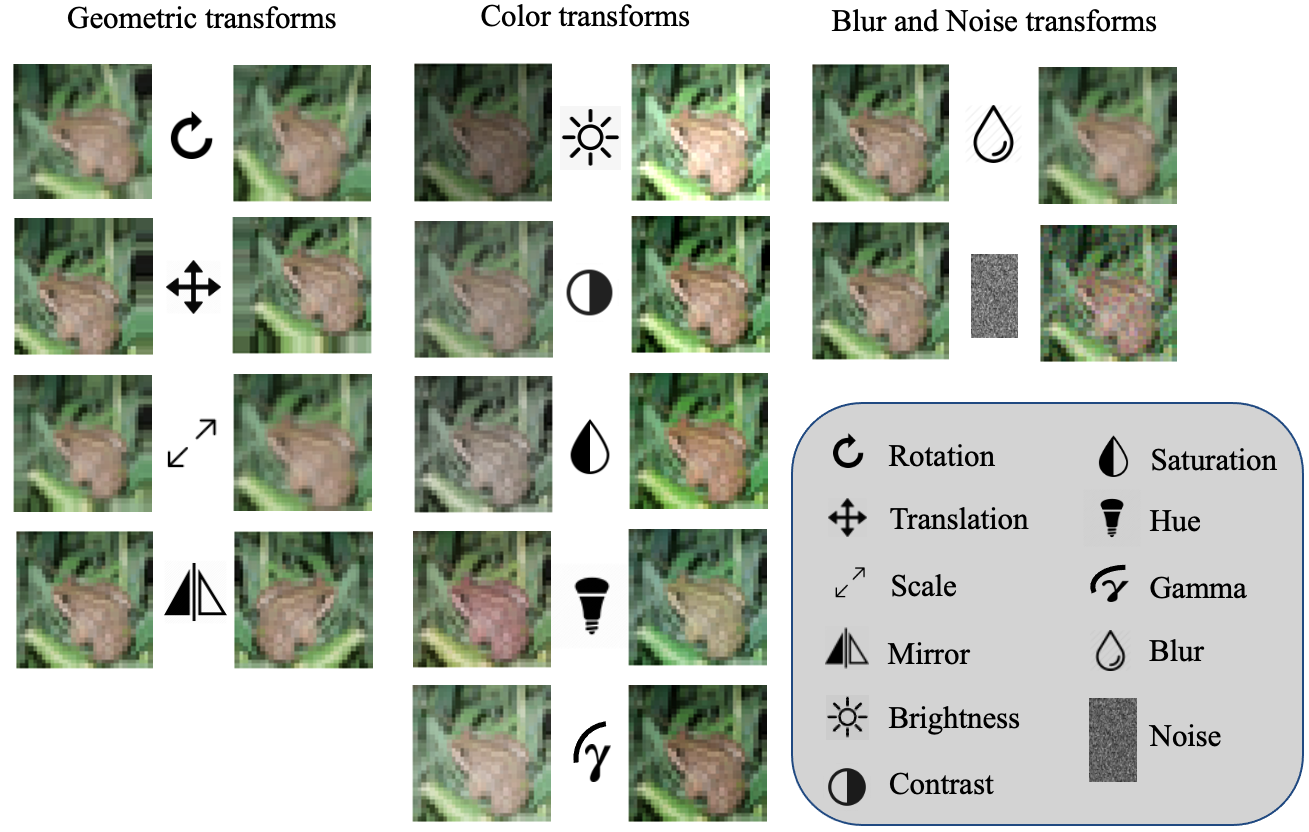}
\caption{All the transforms used for the test-time augmentation (TTA). The left column illustrates the geometric transforms: Rotation, translation, scaling, and horizontal flips (not used on SVHN). The middle column illustrates the color transforms: Brightness, contrast, saturation, hue, and gamma. The right column illustrates a Gaussian blur and an addition of Gaussian white noise. All the above transforms are randomized to generate $N$ TTAs samples.}
\label{fig:transformations}
\vspace{-0.1in}
\end{figure}

We turn now to present our approach. We start by describing how the TTAs are generated prior to feeding them to the DNN. Then we show how the output logits are utilized to train the random forest classifier.

\subsection{Test-time augmentations}
\label{sec:test_time_augmentations}
We hypothesize that even if the adversary succeeds to attack a specific image, the close neighborhood around the image still holds enough information for reverting the predicted (wrong) label back to the correct label. To that end, for each image we generate $N$ TTAs, using a variety of color, geometrical, blur, and noise transforms (see \cref{fig:transformations}).

The color transforms include: Brightness, contrast, saturation, hue, and gamma; the geometric transforms include: Rotation, translation, scaling, and horizontal flipping; the blur transform convolutes the image with a 2D Gaussian kernel $G_{2D}(u, v; \sigma_b)$ where $\sigma_b$ is uniformly distributed for every TTA image between $0.001$ and a positive constant: $\sigma_b \sim \textit{U}(0.001, \sigma_{bmax})$.
The noise transform adds a white Gaussian noise to the image, $x_t = x + n$, where $n$ is sampled from: $n \sim \textit{N}(0, \sigma)$.
and the standard deviation $\sigma$ is uniformly distributed for every TTA image between $0$ and a positive constant $\sigma_{max}$: $\sigma \sim \textit{U}(0, \sigma_{max})$.

All the transforms including their parameters are randomized in test time. More details on the transforms definitions and parameters distributions appear in the Appendix. We chose to apply these transforms because they were shown to improve the classification accuracy significantly in self-supervised and semi-supervised learning \cite{SimCLR}. Similarly to them, all the transforms parameters were chosen to alter the image until a human struggles to perceive the images on the dataset.
We also added the Gamma transform since it showed small improvement (data not shown).

\subsection{TTA classifier}
\label{sec:tta_classifier}
We generate $N$ randomized TTAs and feed them to the DNN (Figure~\ref{fig:arf_flow}), we then collect their logits output ($N$ logits vectors). Formally, for the input TTAs $\{x_t[i]\}_{i\in[0, N-1]}$, the DNN outputs are $\{l[i, c]\}_{i\in[0, N-1]}^{c\in[0, \text{\#classes}-1]}$, where $l[i, c]$ is the logit corresponding to class $c$ of the input image $x_t[i]$.

When using only the TTAs for making the prediction, the inferred label is a simple argmax of the logits summation:
\begin{equation}
c_{pred} = \argmax_c \sum_{i=0}^{N-1} l[i, c].
\label{eq:tta_pred}
\end{equation}

\subsection{ARF classifier}
We split the official test set into two: \textit{test} and \textit{test-val} (see \cref{sec:exp_setup} - Random forest training). Let $TV$ be the \textit{test} size. The augmented random forest (ARF)  employs the aforementioned DNN logits of \textit{test} to train a random forest classifier. We generate $N$ TTAs for the normal (unperturbed) images in \textit{test} and additional $10N$ TTAs for adversarial images of \textit{test} generated using ten generic (non-adapted) adversarial attacks we utilized (see \cref{sec:exp_setup} - Adversarial attacks), we denote them by $\{x_t[k, i]\}$, $\{x'_t[k, i]\}_{k\in[0, TV-1], i\in[0, 10N-1]}$, respectively.

The above TTAs are fed to the DNN and their logits output for the normal and adversarial images are denoted as $\{l[k, i, c]\}$, $\{l'[k, i, c]\}_{k\in[0, TV-1], i\in[0, 10N-1]}^{c\in[0, \text{\#classes}-1]}$, respectively, or $\{l[k]\}$ and $\{l'[k]\}$ in short for clarity. We then fit the random forest classifier using the pairs $\{l[k], y[k]\}\bigcup\{l'[k], y[k]\}$ where $y[k]$ is the true label of the image $x[k]$, i.e., it learns to infer correct labels both from regular and adversarial logits.

The random forest training procedure needs to be carried out only once. For every new (unseen) image we generate TTAs, obtain their logits $l[i, c]$ (as in Section~\ref{sec:tta_classifier}) and feed them to the random forest classifier to predict the class label.


\subsection{Adversarial attacks}
\label{sec:adversarial_attacks}
To inspect our defense against adversarial images, we employed extensive and diverse attacks in a variety of threat models, and then evaluate them using our ARF classifier and compare them to the robustness obtained using equivalent adversarially trained TRADES/VAT networks, and to an ensemble of networks.

\noindent\textbf{Black-box.}
A threat model where the adversary has access only to the DNN output, but neither to the DNN nor to the random forest classifier. In this setup we apply targeted Boundary \cite{Boundary2018} and untargeted Square \cite{SquareAttack} attacks on the DNN. It is the least interesting setting, but used for attack diversity.

\noindent\textbf{Gray-box.}
In the Gray-box threat model the adversary has full access the the DNN parameters, but is oblivious to the pre-processing (transformation) and post-processing (random forest) defenses. In this setup we apply the FGSM, JSMA, PGD, Deepfool, and CW attacks on the DNN. All these attack are targeted except of the Deepfool.

\noindent\textbf{Adaptive black-box.}
In this threat model the adversary does not have any information on the DNN and random forest parameters. They can only query the final output (predicted label) of the random forest classifier and perturb the input image without any gradients knowledge. In this threat model we use the Square attack, which was shown to be much more efficient compared to the popular Boundary attack, and achieved SOTA results, even compared to white-box attacks. Also, we set an untargeted setting since this attack excels on it \cite{SquareAttack}.

\noindent\textbf{Adaptive Gray-box.}
In this threat model the adversary has access to the DNN's parameters and has full knowledge about the distributions of the test time augmentations. The adversary is still oblivious to the post-processing (random forest). We formulate two adaptive attacks for this setting:

\noindent\textbf{1) A-FGSM}: 
This attack applies the FGSM attack on every one of the generated TTAs in $\{x_t[i]\}_{i\in[0, N-1]}$. All the gradients are then averaged and the mean gradient map is added to the original input image. Formally, we define $X_t$ to be the distribution of the generated TTA transforms on an image $x$. Given a loss function $J(x, y; w)$ where $x$ is the input image, $y$ is the adversarial label and $w$ are the DNN weights, the A-FGSM creates an adversarial image $x'$ by:
\begin{equation}
\label{eq:FGSM_WB}
\begin{split}
x' &= x + \mathbb{E}_{x_t \sim X_t}\big[\epsilon \cdot sign\big(\nabla_{x_t} J(x_t, y; w)\big)\big] \\
   &= x + \frac{\epsilon}{N} \sum_{i=0}^{N-1} sign\big(\nabla_{x_t[i]} J(x_t[i], y; w)\big).
\end{split}
\end{equation}

\noindent\textbf{2) A-PGD}: 
Similarly to the gradient averaging shown for A-FGSM, this adaptive attack employs PGD but in every iteration it projects the adversarial perturbations after the addition of the \emph{averaged} TTAs gradients. Formally, let $\delta_k$ be the perturbation added to input image x in step $k$ and $\alpha$ be the perturbation step size. The vanilla PGD attack is:
\begin{align*}
\delta_{k+1} = \mathcal{P}\bigg[\delta_k + \alpha \cdot sign\big(\nabla_{\delta_k}J(x + \delta_k, y; w)\big)\bigg],
\end{align*}
where $\mathcal{P}$ is the projection operator, clipping every perturbation inside a ball of interest defined by a given norm $\norm{\cdot}$ (we use $L_\infty$). For a general norm $\norm{\cdot}$ it simply reads as:
\begin{align*}
\mathcal{P}(\delta) = 
\begin{dcases}
\frac{\delta}{||\delta||}\epsilon,     & \text{if } ||\delta|| > \epsilon\\
~~~\delta,              & \text{otherwise.}
\end{dcases}
\end{align*}

Our adaptive PGD attack is defined as:
\begin{equation}
\label{eq:PGD_WB}
\begin{split}
\delta_{k+1} &= \mathcal{P}\bigg[\delta_k + \mathbb{E}_{x_t \sim X_t}\big[\alpha \cdot sign\big(\nabla_{\delta_k}J(x_t + \delta_k, y; w)\big)\big]\bigg] \\
&=\mathcal{P}\bigg[\delta_k + \alpha \sum_{i=0}^{N-1} sign\big(\nabla_{\delta_k}J(x_t[i] + \delta_k, y; w)\big)\bigg],
\end{split}
\end{equation}
i.e., similar to PGD but averaging the gradients over the batch of augmentations. 
We initialize $\delta_0$ as a zero gradient map and set the generated adversarial image as $x' = x + \delta_N$.


\noindent\textbf{Adaptive white-box}
In this threat model the adversary has full knowledge about the DNN, the distribution of the transformations, and the random forest's parameters. The adversary knows everything about our defense method, including the training data (logits) used to fit the DNN and the random forest classifer. In this harsh settings we employ the BPDA attack \cite{Athalye2018ObfuscatedGG}.
To estimate the random forest gradients, we use knowledge distillation and mimic the random forest classifier (excluding the DNN) to an MLP with 6 layers \cite{DBLP:journals/corr/PapernotMG16} (see Appendix for full details and MLP architecture). Next, we concatenate the DNN + MLP in tandem to a substitute model. This substitute model is used to generate the BPDA adversarial examples.

\section{Experimental setup}
\label{sec:exp_setup}

We turn to detail the datasets we used, the DNN and random forest training, and inference computation time. Our hardware setup is thoroughly detailed in the Appendix.

\noindent\textbf{Datasets.}
We perform our tests on four datasets: CIFAR-10, CIFAR-100 \cite{CIFAR}, SVHN \cite{SVHN} and Tiny ImageNet \cite{TinyImageNet}.

\noindent\textbf{DNN training.}
We randomly split the training set of all the datasets into two subsets, \textit{train} and \textit{train-val}. The former is used to back-prop gradients from the loss to the inputs and train the DNN, whereas the latter is used for metric calculation to decay the learning rate. The size of the \textit{train-val} set is chosen to be 5\% of the official training set.

We trained three Resnet architectures \cite{RESNET}, Resnet-34, Resnet-50, and Resnet-101, with global average pooling layer before the embedding space. The embedding vector was multiplied by a fully connected layer for the logits calculation. We trained CIFAR-10, CIFAR-100, SVHN, and Tiny ImageNet with 300, 300, 200, and 300 epochs, respectively; For the TRADES method (adversarial training) we trained them with 100, 100, 100, and 300 epochs, respectively, since we observed that fewer epochs obtain higher adversarial accuracy with TRADES (see Appendix).
All TRADES adversarial robust networks used $1/\lambda=1$, $\epsilon=0.031$, $\alpha=0.007$ ($\epsilon$ step size), on $L_\infty$ norm to match the settings in \cite{TRADES} for fair comparison. The VAT adversarial networks were also trained using $\epsilon$=0.031, with $\alpha=1$ and $\epsilon=1, 1, 3, 1$ for CIFAR-10, CIFAR-100, SVHN, and Tiny ImageNet, respectively \cite{Virtual_Adversarial_Training}.

We use an $L_2$ weight decay regularization of 0.0001 in all our DNN training, a stochastic gradient decent optimizer with momentum 0.9 with Nesterov updates, and a batch size of 100. The training starts with a learning rate of 0.1, which decreases by a factor of 0.9 after 3 epochs of no improvement on the \textit{train-val} accuracy (2 epochs for SVHN).

\noindent\textbf{Random forest training.}
We split the test set of all the datasets into two subsets, \textit{test} and \textit{test-val}. The \textit{test-val} size is 2500, and the \textit{test} is the official test set without these 2500 samples, i.e., 7500, 7500, 23500, and 7500 for CIFAR-10, CIFAR-100, SVHN, and Tiny ImageNet, respectively.
The only exception is the Boundary attack. Due to long processing time, we selected for it only 750, 250 samples from \textit{test}, \textit{test-val} , respectively.
The random forest classifier was trained with 1000 trees, using the Gini impurity criterion. The training time was 71 seconds and was done only once for all the normal/adversarial images on the \textit{test} subset.

\noindent\textbf{Adversarial attacks.}
The adversarial attacks detailed in \cref{sec:adversarial_attacks} were set to the following norms and powers: 
(1) FGSM$^1$: ($L_\infty, \epsilon=0.01$); 
(2) FGSM$^2$: ($L_\infty, \epsilon=0.031$); 
(3) JSMA: ($L_0, \gamma=0.01$); 
(4) PGD$^1$: ($L_\infty, \epsilon=0.01$);
(5) PGD$^2$: ($L_\infty, \epsilon=0.031$);
(6) Deepfool: ($L_2, \epsilon$ is unconstrained);
(7) CW$_{L_2}$: ($L_2, \epsilon$ is unconstrained);
(8) CW$_{L_\infty}$: ($L_\infty, \epsilon=0.031$);
(9) Square: ($L_\infty, \epsilon=0.031$); and
(10) Boundary: ($L_2, \epsilon$ is unconstrained).

Our PGD attacks were applied with a step size of $\alpha=0.003$ with 100 iterations. The above attacks were selected due to their norm diversity, effectiveness, and popularity. Many attacks employ $\epsilon=0.031$ to match the settings in the TRADES baseline \cite{TRADES}, which is the current SOTA.

We also apply the following adaptive attacks detailed in Sections~\ref{sec:adversarial_attacks}: 
(i) A-FGSM($L_\infty, \epsilon=0.031$);
(ii) A-PGD($L_\infty, \epsilon=0.031$);
(iii) A-Square ($L_\infty, \epsilon=0.031$); and
(iv) BPDA: BPDA($L_\infty, \epsilon=0.031$). 
A-FGSM, A-PGD, A-Square, and BPDA were set with $N=256$ generated TTAs. A-PGD, A-Square and BPDA are very time consuming and thus were set with only 10 iterations; Therefore, we set their step size to $\alpha=0.007$.

\noindent\textbf{Testing.}
All the metrics we show in this work were calculated on the \textit{test-val} subset. 
For both TTA and ARF we set $N=256$ unless stated otherwise, i.e., we generate 256 TTAs in inference time, which allows the models to run in a single forward pass on the GPU. The majority of the computation time is devoted to the TTAs generation, which is done on the CPU and takes 3.32 $\pm$ 0.33 seconds for a single Tiny ImageNet image (calculated over 20 runs). The DNN and random forest forward pass times are negligible - 250 ms and 4 ms, respectively.

\section{Results}
We evaluate the performance of ARF on adversarial attacks and compare it to other robust methods. We also present ablation studies conducted to improve the model performance and computation time. Lastly, we show accuracies on adaptive black-box and white-box attacks.
Alternative simple machine learning classifiers such as logistic regression and SVM were found to be inferior to random forest; This comparison appears in the Appendix. 

\noindent {\bf Transferability.} While here we train and test ARF using all non-adaptive attacks (black-box and gray-box), in the Appendix. we demonstrate an excellent generalization to unseen, non-adaptive attacks.

\begin{table*}[ht!]
\centering
\resizebox{1.8\columnwidth}{!}{
    \begin{tabular}{cc|ccccccccccc}
    \toprule
    Dataset & Method & Normal & FGSM$^1$ & FGSM$^2$ & JSMA & PGD$^1$ & PGD$^2$ & Deepfool & CW$_{L_2}$ & CW$_{L_\infty}$ & Square & Boundary \\
    \hline
    \rowcolor{Gray}
    \multirow{7}{*}{CIFAR-10} & Plain & 94.92 & 68.52 & 55.28 & 68.68 & 13.72 & 0.00 & 4.00 & 3.12 & 23.44 & 59.36 & 18.40 \\ 
    \rowcolor{Gray}
    & Ensemble & 96.04 & 82.00 & 64.20 & 84.60 & 86.68 & 48.64 & 86.96 & 83.64 & 78.80 & 89.48 & 96.00 \\
    & TRADES   & 86.64 & 85.04 & 75.80 & 69.88 & 85.12 & 71.84 & 7.68 & 0.56 & 78.24 & 80.92 & 22.80 \\
    & VAT      & \textbf{94.00} & 82.68 & 70.36 & 80.48 & 82.12 & 20.08 & 4.04 & 4.24 & 49.80 & 81.32 & 15.20 \\
    & TTA      & 91.68 & 82.48 & 68.76 & 84.84 & 87.04 & 72.76 & 83.16 & 82.24 & 81.4 & 85.32 & 88.80 \\
    & ARF      & 93.76 & 83.72 & 70.20 & 85.28 & 90.32 & 77.88 & 87.36 & 84.36 & 85.64 & 87.84 & \textbf{91.20} \\
    & TRADES + ARF & 84.28 & 82.56 & 76.72 & 79.64 & 82.56 & 76.24 & 69.40 & 68.00 & 80.48 & 80.56 & 81.20 \\
    & VAT + ARF    & 92.60 & \textbf{89.44} & \textbf{81.24} & \textbf{90.60} & \textbf{90.28} & \textbf{82.36} & \textbf{87.72} & \textbf{85.12} & \textbf{88.92} & \textbf{89.00} & 90.80 \\
    \hline
    \rowcolor{Gray}
    \multirow{7}{*}{CIFAR-100} & Plain & 74.32 & 28.96 & 13.84 & 43.84 & 22.52 & 0.28 & 9.20 & 15.84 & 47.76 & 28.64 & 30.40 \\
    \rowcolor{Gray}
    & Ensemble & 78.04 & 58.20 & 29.16 & 52.48 & 69.64 & 33.28 & 76.60 & 51.08 & 68.68 & 65.36 & 74.40 \\
    & TRADES   & 53.36 & 51.80 & 41.52 & 46.84 & 52.88 & 46.44 & 10.88 & 5.28  & 51.04 & 45.96 & 24.00 \\
    & VAT      & 70.92 & 52.56 & 28.80 & 59.88 & 63.00 & 15.20 & 10.00 & 11.20 & 44.36 & 54.36 & 20.80 \\
    & TTA      & 70.76 & 52.24 & 28.80 & 56.36 & 62.92 & 42.08 & 65.92 & 46.72 & 62.60 & 56.28 & 65.20 \\
    & ARF      & \textbf{71.52} & 54.20 & 30.80 & 58.08 & \textbf{66.72} & 47.60 & \textbf{68.12} & 49.36 & \textbf{64.44} & 60.40 & \textbf{68.00} \\
    & TRADES + ARF & 49.48 & 49.04 & 44.16 & 48.04 & 48.80 & 46.28 & 45.04 & 36.56 & 49.96 & 47.48 & 43.60 \\
    & VAT + ARF    & 68.96 & \textbf{63.52} & \textbf{48.20} & \textbf{65.76} & 66.24 & \textbf{59.92} & 65.56 & \textbf{55.76} & 62.72 & \textbf{64.04} & 61.60 \\
    \hline
    \rowcolor{Gray}
    \multirow{7}{*}{SVHN} & Plain & 97.36 & 80.56 & 66.24 & 49.64 & 52.32 & 2.04 & 2.84 & 4.80 & 28.96 & 66.96 & 15.60 \\
    \rowcolor{Gray}
    & Ensemble & 98.12 & 89.52 & 74.84 & 85.36 & 92.80 & 71.20 & 71.88 & 79.76 & 83.04 & 93.52 & 97.20 \\
    & TRADES   & 92.48 & 90.60 & 81.20 & 39.44 & 90.28 & 70.88 & 5.08  & 0.72  & 82.36 & 83.84 & 19.20 \\
    & VAT      & 94.44 & 89.00 & 83.72 & 65.48 & 85.16 & 43.28 & 4.16  & 23.60 & 64.76 & 87.04 & 18.80 \\
    & TTA      & 97.08 & 87.16 & 73.76 & 86.36 & 89.68 & 61.32 & 66.84 & 80.08 & 80.64 & 92.24 & 95.20 \\
    & ARF      & \textbf{96.92} & 87.96 & 75.24 & 87.00 & 90.04 & 66.16 & 68.84 & 80.40 & 81.28 & 92.16 & \textbf{96.00} \\
    & TRADES + ARF & 92.44 & 90.96 & 82.20 & 78.80 & 91.16 & 79.96 & 62.48 & 48.40 & 85.84 & 88.40 & 88.00 \\
    & VAT + ARF    & 95.64 & \textbf{93.72} & \textbf{86.92} & \textbf{89.84} & \textbf{93.76} & \textbf{81.88} & \textbf{92.08} & \textbf{85.68} & \textbf{90.28} & \textbf{93.24} & 91.60 \\
    \hline
    \rowcolor{Gray}
    \multirow{7}{*}{Tiny ImageNet} & Plain & 59.24 & 25.48 & 9.92 & 28.88 & 30.72 & 0.40 & 10.08 & 16.24 & 34.76 & 28.04 & 19.60 \\
    \rowcolor{Gray}
    & Ensemble & 67.12 & 58.32 & 28.64 & 52.96 & 63.92 & 46.24 & 66.56 & 54.64 & 60.08 & 60.80 & 58.40 \\
    & TRADES   & 44.44 & 42.32 & 31.64 & 35.32 & 43.72 & 38.44 & 9.16 & 5.52 & 41.88 & 38.76 & 23.20 \\
    & VAT      & \textbf{54.68} & \textbf{47.84} & 24.52 & 44.92 & \textbf{52.36} & 31.80 & 9.92 & 6.64 & \textbf{46.24} & \textbf{46.00} & 23.20 \\
    & TTA      & 52.48 & 37.12 & 17.36 & 39.76 & 43.84 & 27.52 & 46.24 & 35.08 & 42.96 & 40.84 & 37.60 \\
    & ARF      & 53.36 & 40.76 & 21.52 & 43.76 & 48.88 & 33.48 & \textbf{48.88} & \textbf{40.04} & 45.80 & 45.92 & \textbf{42.80} \\
    & TRADES + ARF & 39.24 & 37.68 & 33.44 & 36.20 & 38.36 & 35.20 & 35.12 & 27.80 & 38.72 & 36.80 & 32.00 \\
    & VAT + ARF    & 47.92 & 45.52 & \textbf{36.28} & \textbf{45.76} & 46.32 & \textbf{41.96} & 43.44 & 34.84 & \textbf{46.24} & 44.84 & 39.20 \\
    \bottomrule
    \end{tabular}
}
\caption{Comparison of accuracies (\%) for various classifiers on the non-adaptive attacks. The tested attacks and datasets are detailed in Section~\ref{sec:exp_setup}. We boldface the best results. Ensemble is presented just as a reference as it has an unfair advantage (as explained in the text).}.
\label{table:robustness_scores_resnet34}
\vspace{-0.18in}
\end{table*}

\subsection{Adversarial Robustness}
Table~\ref{table:robustness_scores_resnet34} shows the accuracy on the normal (not attacked) and the adversarial accuracies obtained for all the non-adaptive attacks (black-box and gray-box) we employed (see Section~\ref{sec:exp_setup}), on Resnet-34. Tables for Resnet-50 and Resnet-101 are shown in the Appendix.

"Plain" corresponds to the non-robust, simple DNN accuracy, without any adversarial defense method. "Ensemble" uses nine different DNNs with the same architecture and the predicted label is a majority voting amongst them. It should be emphasized that the adversary does not have access to any of these nine models. Thus, it has an unfair advantage. TTA classifier is applied on the DNN alone (w/o random forest), and ARF classifier is evaluated in two setups, one is applied on a regular (non adversarially robust) DNN, and second is combined with an adversarially robust DNN trained with VAT or TRADES. 


We notice that for every dataset both TTA and ARF classifiers achieve higher robustness accuracy than both VAT and TRADES on JSMA, Deepfool, and CW$_{L_2}$. This is not surprising because TRADES employed a regularization term on a ball with an $L_\infty$ norm and these attacks use other norms. In addition, VAT regularizes logits distribution smoothness within the image's local surrounding, which is problematic where the norm is unconstrained in $L_\infty$. Nonetheless, it is not trivial that the simple TTA classifier surpasses TRADES' accuracy on these attacks.

In the vast majority of the attacks and datasets, the ARF classifier outperforms the VAT networks. However, when combined together they usually achieve the highest adversarial robustness accuracy. This robust accuracy trumps even the ensemble score, except for Tiny Imagenet.




Lastly, we observe that the normal accuracy obtained by ARF is much better than the one of TRADES, and is comparable to the normal accuracy obtained by VAT. ARF scores almost as the "plain" classifier for normal images on CIFAR-10, CIFAR-100, and SVHN.


\subsection{Ablation Studies}
We conducted two ablation studies to understand better and optimize our ARF and TTA classifiers.

\noindent\textbf{ARF classifier ablation.}
\label{sec:ARF_param_ablation}
We tested three parameters governing the ARF accuracy:
\begin{enumerate}[leftmargin=*]
\item \emph{Features:} The inputs to the random forest classifier. We used three candidates: The DNN's logits, the DNN's probabilities (softmax over logits), and the embedding vectors in the DNN penultimate layer.
\item \emph{Gaussian noise power:} We checked three different noise filters with max standard deviation ($\sigma_{max}$) of $0$, $0.005$, and $0.0125$. The $0$ value is equivalent to no noise.
\item \emph{Strength of transforms:} We tested two sets of transforms from the transforms in \cref{fig:transformations}: \textit{soft} vs \textit{hard}. The \textit{soft} transforms span over shorter parameter intervals. For example, the \textit{hard} brightness transform randomizes a brightness factor in the interval $\textit{U}(0.6, 1.4)$ whereas the \textit{soft} transform randomizes it in $\textit{U}(0.8, 1.2)$. The full interval sets of the soft and hard transforms are listed in the Appendix.
\end{enumerate}

\begin{table}
\centering
\resizebox{0.95\columnwidth}{!}{
    \begin{tabular}{ccc|cc}
    \toprule
    \multirow{2}{*}{Features} & \multirow{2}{*}{Transforms} & \multirow{2}{*}{$\sigma_{max}$} & \multicolumn{2}{c}{Accuracy (\%)} \\
    & & & $\mathbb{A}_{norm}$ & $\mathbb{A}_{adv}$ \\
    \hline
    Logits & \textit{soft} & 0      & \textbf{94.24} & 83.56 \\
    Logits & \textit{soft} & 0.005  & 94.20 & 83.72 \\
    Logits & \textit{soft} & 0.0125 & 93.88 & 83.96 \\
    Logits & \textit{hard} & 0      & 93.72 & 84.64 \\
    Logits & \textit{hard} & 0.005  & 93.80 & \textbf{85.00} \\
    Logits & \textit{hard} & 0.0125 & 93.08 & 84.96 \\
    \hdashline
    Probs & \textit{soft} & 0      &  94.16 & 83.36 \\
    Probs & \textit{soft} & 0.005  &  94.04 & 83.64 \\
    Probs & \textit{soft} & 0.0125 &  93.80 & 84.00 \\
    Probs & \textit{hard} & 0      &  93.60 & 84.52 \\
    Probs & \textit{hard} & 0.005  &  93.80 & 84.72 \\
    Probs & \textit{hard} & 0.0125 &  93.00 & 84.96 \\
    \hdashline
    Embeddings & \textit{soft} & 0      & 94.16 & 83.56 \\
    Embeddings & \textit{soft} & 0.005  & 93.96 & 83.64 \\
    Embeddings & \textit{soft} & 0.0125 & 93.56 & 83.88 \\
    Embeddings & \textit{hard} & 0      & 93.68 & 84.88 \\
    Embeddings & \textit{hard} & 0.005  & 93.60 & 84.80 \\
    Embeddings & \textit{hard} & 0.0125 & 93.04 & 84.92 \\
    \bottomrule
    \end{tabular}
}
\caption{Ablation study on 3 parameters used for ARF. 1) Random forest input features: Logits, softmax probabilities, and DNN embeddings. 2) Randomization level of transforms: \textit{hard} for a larger randomization range (coarse transforms) and \textit{soft} for a smaller range (mellow transforms). 3) Noise transform max power ($\sigma_{max}$).}
\label{table:ablation_params}
\vspace{-0.18in}
\end{table}

Table~\ref{table:ablation_params} shows the normal and adversarial accuracies ($\mathbb{A}_{norm}$ and $\mathbb{A}_{adv}$) on CIFAR-10, trained by Resnet-34, attacked by CW$_{L_2}$ and evaluated using ARF with $N=1000$. The highest adversarial accuracy was obtained for logits vectors, \textit{hard} transforms, and $\sigma_{max}=0.005$. Thus these were the parameters we used in this work.
It is interesting to point out that the best normal accuracy was obtained for \textit{soft} transforms with $\sigma_{max}=0$ (for all features). This observation conforms with the high normal accuracy presented in Table~\ref{table:robustness_scores_resnet34}, as the plain DNN does not apply any transform.

\begin{figure}[t]
\centering
\includegraphics[width=\linewidth]{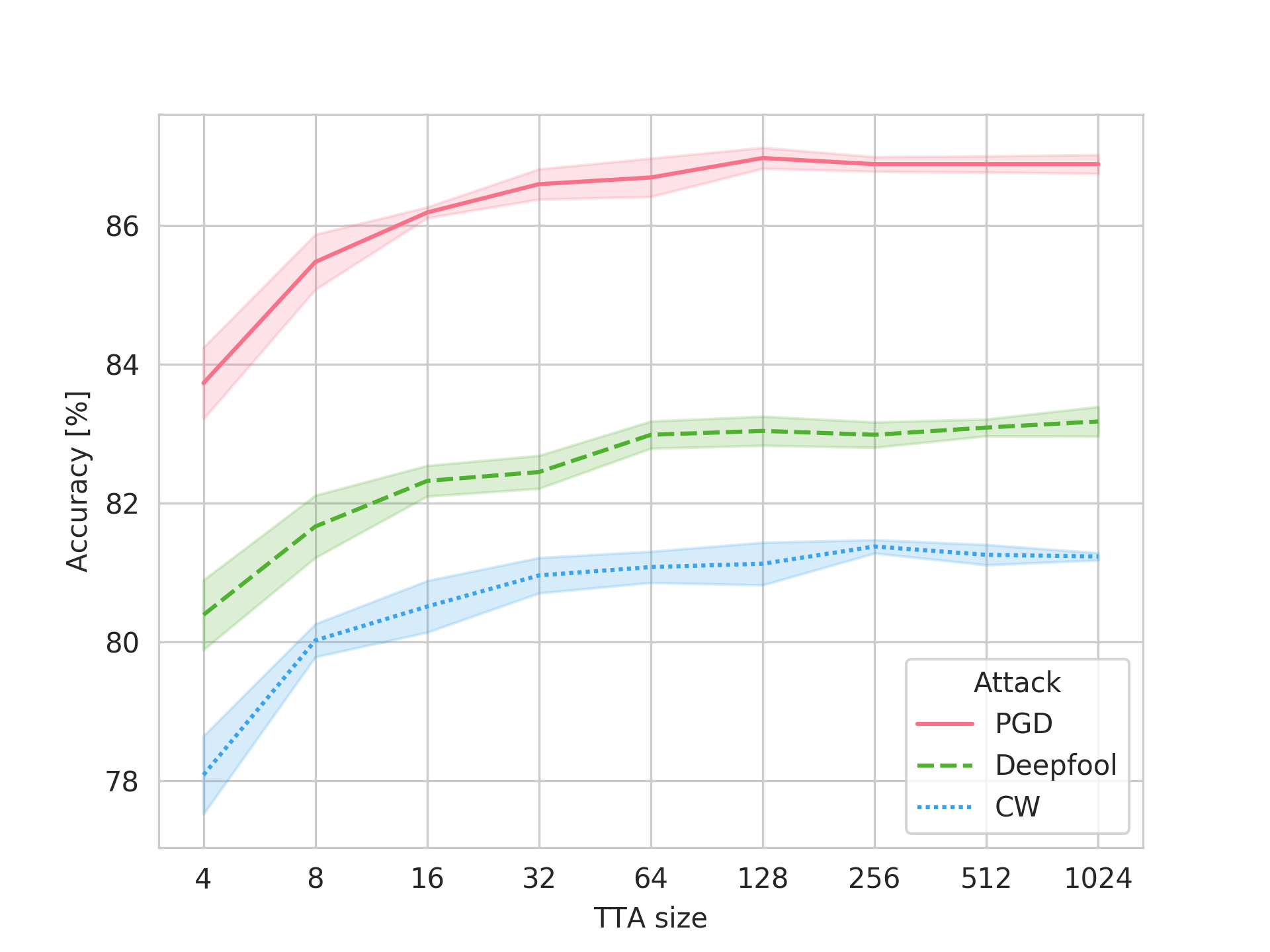}
\caption{Ablation study on the number of generated TTAs ($N$). We calculate the adversarial accuracies on CIFAR-10 for three attacks as a function of $N$ (logarithmic scale).}
\label{fig/tta_size_ablation_cifar10}
\vspace{-0.18in}
\end{figure}

\noindent \textbf{TTA size ablation.}
The computational bottleneck in our TTA and ARF classifiers is the generation of the $N$ TTAs. Using $N=1000$ images as done for Table~\ref{table:ablation_params} requires a long computation time so we searched for the minimal $N$, which achieves sufficient adversarial robustness. Figure~\ref{fig/tta_size_ablation_cifar10} shows the adversarial accuracy on CIFAR-10 for three selected attacks: PGD$^1$, Deepfool, and CW$_{L_2}$ in a logarithmic scale. The width of each line corresponds to the measured standard deviation of five repeated experiments. We select $N=256$ for our experiments since it achieves good robustness with very high confidence (narrow interval). Ablation of the TTA size on CIFAR-100 and SVHN is presented in the Appendix.

\subsection{Adaptive Attacks (Limitation)}
\label{sec:adaptive_attack_results}
Table~\ref{table:adaptive_attacks} shows the adversarial accuracies for different robust classifiers for all the adaptive attacks shown in Section~\ref{sec:adversarial_attacks}. For easy comparison, the performance on the corresponded non-adaptive attack is shown above each accuracy result. We observe that our ARF defense is robust against the black-box adaptive attack, but fails when attacked with an adaptive gray-box or white-box attack. For example, the white-box BPDA attack decreases the ARF accuracy on CIFAR-10 to 8.8\%.

The VAT+ARF combination demonstrates SOTA robustness for all non-adaptive attacks, however, the vanilla TRADES or VAT perform better on adaptive attacks. We observe that the combination of TRADES+ARF is usually preferred against adaptive attack, obtaining high robustness for these attacks for all datasets. Note that although the performance of ARF degrades significantly when the BPDA harsh attack is applied, when combined with adversarial training this degradation is minimal.

Our A-PGD attack is much more effective against the ensemble and TTA classifiers, surpassing all other adaptive and non-adaptive attacks by a large margin. Alas, it is not as powerful as the vanilla PGD against plain adversarial robust DNNs (TRADES/VAT). For all the other robust classifiers it achieves comparable results to the strong BPDA attack.

\begin{table}[t]
\centering
\resizebox{\columnwidth}{!}{
    \begin{tabular}{cc|ccccccccc}
    \toprule
    \multirow{2}{*}{Dataset} & \multirow{2}{*}{Attack} & \multirow{2}{*}{Ensemble} & \multirow{2}{*}{TRADES} & \multirow{2}{*}{VAT}  & \multirow{2}{*}{TTA} & \multirow{2}{*}{ARF} & TRADES+ & TRADES+ & VAT+ & VAT+ \\
    & & & & & & & TTA & ARF & TTA & ARF \\

    \hline
    \multirow{7}{*}{CIFAR-10}  & FGSM        & 64.20 & 75.80 &  70.36   &68.76 & 70.28 & 73.72 & 76.72 &80.40&\textbf{81.24}\\
    &                            A-FGSM      & 41.60 & \textbf{79.20} &  68.84   &33.32 & 37.80 & 72.96 & 74.88 &61.88&64.68\\
    &                            PGD         & 48.64 & 71.84 &  20.08   &72.76 & 78.32 & 72.84 & 76.24 &79.28&\textbf{82.36}\\
    &                            A-PGD       & 5.76  & \textbf{77.76} &  54.84   &5.12  & 22.32 & 70.40 & 73.24 &53.20&54.64\\
    &                            Square      & 89.48 & 80.92 &  81.32   &85.32 & 87.84 & 77.44 & 80.56 &87.20&\textbf{89.00}\\
    &                            A-Square    & 90.80  & 89.60 &  \textbf{95.2}   &87.60 & 89.20 & 84.00 & 87.20 &90.80&92.00\\
    &                            BPDA        & 10.40  & \textbf{84.00} &  58.8   &8.40  & 8.80 & 74.80  & 82.80 &57.20&63.60\\
    \hline
    \multirow{7}{*}{CIFAR-100} & FGSM        & 29.16 & 41.52 & 28.80   &28.80 & 30.80 & 43.28 & 44.16 &46.16&\textbf{48.20}\\
    &                            A-FGSM      & 25.04 & \textbf{48.76} & 40.32   &13.20 & 16.60 & 45.88 & 46.44 &34.32&37.80\\
    &                            PGD         & 33.28 & 46.44 & 15.20   &42.08 & 47.60 & 44.96 & 46.28 &56.76&\textbf{59.92}\\
    &                            A-PGD       & 19.24 & \textbf{48.44} & 39.32   &10.16 & 13.24 & 45.96 & 47.28 &40.60&43.16\\
    &                            Square      & 65.36 & 45.96 & 54.36   &56.28 & 60.40 & 47.88 & 47.48 &61.28&\textbf{64.04}\\
    &                            A-Square    & 66.40 & 52.80 & 64.00   &56.00 & 58.00 & 48.40 & 47.20 &63.20&\textbf{64.80}\\
    &                            BPDA        & 23.60 & \textbf{50.40} & 43.60   &12.00 & 12.80 & 44.40 & 46.00 &38.00&42.40\\
    \hline
    \multirow{7}{*}{SVHN}      & FGSM        & 74.84 & 81.20 & 83.72   &73.76 & 75.24 & 80.72 & 82.20 &85.92&\textbf{86.92}\\
    &                            A-FGSM      & 62.44 & \textbf{82.00} & 79.52   &56.92 & 59.80 & 78.44 & 80.96 &79.32&79.68\\
    &                            PGD         & 71.20 & 70.88 & 43.28   &61.32 & 66.16 & 77.64 & 79.96 &80.28&\textbf{81.88}\\
    &                            A-PGD       & 36.04 & \textbf{78.40} & 58.32   &19.92 & 34.28 & 74.68 & 76.60 &64.84&65.36\\
    &                            Square      & 93.52 & 83.84 & 87.04   &92.24 & 92.16 & 83.88 & 88.40 &91.88&\textbf{93.24}\\
    &                            A-Square    & 96.80 & 92.40 & 92.40   &\textbf{96.40} & 95.60 & 88.40 & 91.60 &93.20&94.00\\
    &                            BPDA        & 45.60 & \textbf{75.60} & 62.80   &32.80 & 34.80 & 66.80 & 72.80 &63.60&66.40\\
    \hline
    \multirow{7}{*}{Tiny ImageNet} & FGSM    & 28.64 & 31.64 & 24.52   &17.36 & 21.52 & 29.60 & 33.44 &31.24&\textbf{36.28}\\
    &                                A-FGSM  & 30.04 & 40.76 & \textbf{43.12}   &6.96  & 10.40 & 31.44 & 35.24 &28.64&36.16\\
    &                                PGD     & 46.24 & 38.44 & 31.80   &27.52 & 33.48 & 31.56 & 35.20 &34.96&\textbf{41.96}\\
    &                                A-PGD   & 38.16 & 40.00 & \textbf{46.28}   &6.72  & 10.00 & 33.60 & 35.80 &35.00&40.56\\
    &                           Square       & 60.80 & 38.76 & \textbf{46.00}   &40.84 & 45.92 & 34.84 & 36.80 &41.32&44.84\\
    &                           A-Square     & 56.80 & 40.80 & \textbf{49.60}   &40.00 & 45.60 & 31.60 & 38.00 &37.20&38.00\\
    &                           BPDA         & 39.60 & 37.60 & \textbf{43.60}   &11.60 & 15.20 & 31.20 & 35.60 &30.00&36.40\\
    \bottomrule
    \end{tabular}
}
\caption{Adversarial accuracies (\%) for various robust classifiers on adaptive attacks: A-Square (black-box), A-FGSM and A-PGD (gray-box) and BPDA (white-box), and their non-adaptive correspondents. FGSM$^2$ and PGD$^2$ are abbreviated to FGSM and PGD for clarity. TTA and ARF methods can maintain robustness only when combined with an adversarially trained DNN. Ensemble is presented just as a reference as it has an unfair advantage (see text).}
\vspace{-0.18in}
\label{table:adaptive_attacks}
\end{table}

\section{Conclusions}
This work proposes a simple, fast, and easy to use method to classify adversarial images, named ARF. Our approach is applied on pretrained DNNs without the need to carry out adversarial training or updating the model's parameters. ARF first generates many test-time augmentations, applying a wide variety of random color, geometrical, blur and noise transforms on the input image, and feeds these augmentations to a pretrained DNN.
Then it collects the DNN's logits and feeds them to a vanilla random forest classifier which yields SOTA robust classification when combined with an adversarially trained DNN (VAT). This improvement in robustness comes at the cost of training the random forest model (only once). We tested ARF with a variety of attacks, where some of them were especially designed against ARF. One of them, A-PGD, which we proposed, is of interest by itself as it is very effective against DNN ensemble while not having access to any of its networks.

ARF can be incorporated to work with any machine learning classifier and was shown to perform well even under the harsh white-box threat model when combined with TRADES.
Note that white-box setting assumes full knowledge about our defense parameters, which can be easily changed by quickly re-training the simple ARF model. Thus, hiding the ARF model can be considered as holding a secret key for "security through obscurity" \cite{OnEvaluatingAdvRobustness2019,Kerckhoffs1883}. In addition, defending against new adaptive attacks is feasible by including them into the ARF fitting. Therefore, the use of ARF should be favored over adversarial training alone (although in the white-box setting tailored to ARF it was better alone) as in the non white-box setting ARF leads to a significant improvement. We believe that integrating ARF within the adversarial training can further boost the robustness as was shown for data augmentations in a very recent work \cite{rebuffi2021data}.



{
    \small
    \bibliographystyle{ieee_fullname}
    \bibliography{my_bib}
}

\onecolumn
\appendix
\vspace{-0.5in}
In Appendix~\ref{sec:other_nets} we present adversarial robustness scores of our approach using other architectures: Resnet-50 and Resnet-101, demonstrating that the combination of VAT+ARF achieves SOTA robustness as shown also for Resnet-34 in the main paper.

Appendix~\ref{sec:test_time_aug} defines in detail the image transformations we used to calculate the Test Time Augmentations (TTAs) in the paper, listing their parameter distribution and randomized ordering protocol.

Appendix~\ref{sec:hardware} lists the hardware (CPUs \& GPUs) we used for training our DNNs and random forest classifiers.

Appendix~\ref{sec:adversarial_training} shows several parameter searches we conducted to optimize our baselines (TRADES/VAT) to the datasets and architecture in our experiments.

Appendix~\ref{sec:alternative_cls} shows ARF performance when replacing the random forest classifier with other machine learning models: Logistic regression and SVM.

In Appendix~\ref{sec:transferability} we show that our ARF model is transferable, generalizing very well to new (unseen) attacks.

Appendix~\ref{sec:bpda} provides in depth details on the steps we implemented to utilize the BPDA attack in our experiments, since the non-differential random forest had to be mimicked by a substitute model.

Appendix~\ref{sec:tta_size_ablation} is a continuation to the TTA size ablation study conducted in Section~5.2 in the main paper. Here, we add the same "accuracy vs size" results for CIFAR-100 and SVHN.

Appendix~\ref{sec:L2_distortion} shows the mean $L_2$ distortion for some attacks we used in the paper.

Lastly, Appendix~\ref{sec:visual} displays adversarial images generated by the adaptive white-box BPDA attack against our defense method. We show that albeit BPDA can circumvent our defense for a vanilla DNN (w/o TRADES/VAT), its generated noise can be observed by the naked eye.

\section{Robustness on Resnet-50 and Resnet-101}
\label{sec:other_nets}
\setcounter{table}{0}
\renewcommand{\thetable}{A\arabic{table}}

The main paper shows adversarial robustness results only on Resnet-34. In this section we repeat the results in Section~5.1 also for Resnet-50 and Resnet-101, shown in Table~\ref{table:robustness_scores_resnet50_all} and Table~\ref{table:robustness_scores_resnet101_all}, respectively. We omit the black-box Boundary attack from these experiments because it utilizes thousands of search queries, which is not practical for large DNN architectures. The black-box Square attack is reported since it is fast and efficient.

\begin{table}[h!]
\centering
\begin{adjustwidth}{-0.7cm}{}
    \begin{tabular}{cc|ccccccccccc}
    \toprule
    Dataset & Method & Normal & FGSM$^1$ & FGSM$^2$ & JSMA & PGD$^1$ & PGD$^2$ & Deepfool & CW$_{L2}$ & CW$_{L\infty}$ & Square \\
    \hline
    \rowcolor{Gray}
    \multirow{8}{*}{CIFAR-10} & Plain & 94.80 & 59.12 & 40.00 & 76.24 & 8.16 & 0.00 & 3.44 & 0.08 & 16.00 & 51.08 \\
    \rowcolor{Gray}
    & Ensemble & 95.92 & 78.04 & 53.00 & 86.48 & 79.84 & 22.00 & 93.56 & 85.92 & 79.48 & 86.20 \\
    & TRADES   & 86.56 & 84.40 & 75.44 & 71.72 & 84.52 & 70.68 & 8.36 & 0.24 & 78.40 & 81.52 \\
    & VAT      & \textbf{95.16} & 80.40 & 63.44 & 84.88 & 72.56 & 5.08 & 3.56 & 1.16 & 25.24 & 79.24 \\
    & TTA      & 90.80 & 78.64 & 59.92 & 84.64 & 83.52 & 57.08 & 87.36 & 82.80 & 81.48 & 82.04 \\
    & ARF   & 93.28 & 81.92 & 63.16 & 85.04 & 87.76 & 60.96 & \textbf{91.36} & 86.24 & 86.64 & 85.44 \\
    & TRADES + ARF & 84.16 & 82.20 & 75.96 & 78.64 & 82.00 & 74.60 & 67.48 & 67.16 & 80.16 & 80.20 \\
    & VAT + ARF & 93.24 & \textbf{89.36} & \textbf{77.64} & \textbf{90.76} & \textbf{90.48} & \textbf{78.36} & 89.84 & \textbf{88.00} & \textbf{87.32} & \textbf{88.80} \\
    \hline
    \rowcolor{Gray}
    \multirow{8}{*}{CIFAR-100} & Plain & 74.52 & 26.48 & 11.08 & 47.64 & 16.52 & 0.08 & 9.44 & 8.08 & 36.56 & 24.52 \\
    \rowcolor{Gray}
    & Ensemble & 78.44 & 53.60 & 23.08 & 56.72 & 64.48 & 32.64 & 76.88 & 53.84 & 62.40 & 60.36 \\
    & TRADES & 55.08 & 53.92 & 44.60 & 45.92 & 54.08 & 48.64 & 11.08 & 5.52 & 52.64 & 47.96 \\
    & VAT & \textbf{71.56} & 52.56 & 28.84 & 60.04 & 63.88 & 13.76 & 9.72 & 6.40 & 40.52 & 52.96 \\
    & TTA    & 67.32 & 49.60 & 26.56 & 54.84 & 60.76 & 46.08 & 64.20 & 49.52 & 58.00 & 52.56 \\
    & ARF & 69.24 & 54.08 & 31.40 & 57.40 & 64.88 & 52.28 & \textbf{65.68} & 53.80 & \textbf{62.00} & 56.36 \\
    & TRADES + ARF & 50.68 & 49.80 & 45.20 & 47.04 & 49.32 & 46.64 & 46.56 & 34.60 & 50.20 & 47.84 \\
    & VAT + ARF & 67.28 & \textbf{62.72} & \textbf{48.84} & \textbf{65.60} & \textbf{65.36} & \textbf{59.88} & 64.24 & \textbf{56.16} & 61.92 & \textbf{62.40} \\
    \hline
    \rowcolor{Gray}
    \multirow{8}{*}{SVHN} & Plain & 97.28 & 81.48 & 65.12 & 51.64 & 51.32 & 0.88 & 2.36 & 2.60 & 20.24 & 67.08 \\
    \rowcolor{Gray}
    & Ensemble & 97.88 & 91.32 & 75.28 & 85.48 & 92.92 & 71.28 & 84.80 & 85.72 & 85.92 & 92.72 \\
    & TRADES & 93.72 & 92.92 & \textbf{89.76} & 52.36 & 92.00 & 77.76 & 4.24 & 0.40 & 89.08 & 84.44 \\
    & VAT & 96.52 & 87.96 & 80.92 & 84.88 & 48.28 & 0.20 & 2.24 & 0.36 & 9.36 & 81.68 \\
    & TTA    & \textbf{97.24} & 89.32 & 75.04 & 86.08 & 91.76 & 63.92 & 83.44 & 86.08 & 84.52 & 93.00 \\
    & ARF & 97.16 & 89.68 & 76.72 & 87.40 & 92.48 & 67.44 & 84.28 & 86.56 & 86.36 & 92.72 \\
    & TRADES + ARF & 93.72 & 93.24 & 89.52 & 84.88 & 92.76 & 83.92 & 74.60 & 48.72 & 90.36 & 89.32 \\
    & VAT + ARF & 96.72 & \textbf{95.28} & 87.96 & \textbf{94.24} & \textbf{95.68} & \textbf{87.96} & \textbf{96.28} & \textbf{95.44} & \textbf{95.08} & \textbf{94.76} \\
    \hline
    \rowcolor{Gray}
    \multirow{8}{*}{\footnotesize Tiny ImageNet} & Plain & 64.16 & 25.68 & 11.68 & 32.92 & 30.60 & 0.08 & 9.52 & 17.28 & 41.60 & 28.84 \\
    \rowcolor{Gray}
    & Ensemble & 69.08 & 56.48 & 28.12 & 52.44 & 65.68 & 46.92 & 68.24 & 51.40 & 61.84 & 58.60 \\
    & TRADES & 45.60 & 45.12 & 33.92 & 37.60 & 44.68 & 41.12 & 10.48 & 6.36 & 43.84 & 42.08 \\
    & VAT & \textbf{63.24} & 53.64 & 40.96 & \textbf{56.64} & 23.92 & 0.52 & 9.76 & 18.12 & 54.28 & 41.12 \\
    & TTA    & 51.88 & 36.88 & 19.08 & 40.16 & 43.36 & 30.20 & 45.92 & 33.08 & 42.80 & 38.44 \\
    & ARF & 54.00 & 41.00 & 22.64 & 43.44 & 47.84 & 35.56 & 49.40 & 37.36 & 45.56 & 43.96 \\
    & TRADES + ARF & 37.52 & 36.32 & 31.08 & 35.48 & 36.36 & 33.60 & 33.52 & 25.64 & 37.80 & 36.80 \\
    & VAT + ARF & 56.00 & \textbf{54.04} & \textbf{48.16} & 54.44 & \textbf{56.04} & \textbf{52.84} & \textbf{56.64} & \textbf{48.88} & \textbf{56.76} & \textbf{50.08} \\
    \bottomrule
    \end{tabular}
\end{adjustwidth}
\caption{Comparison of accuracies (\%) for various classifiers on CIFAR-10, CIFAR-100, SVHN, and Tiny ImageNet trained on Resnet-50. All attacks are detailed in Section~4 in the main paper. We boldface the best results. Ensemble is presented just as a reference as it has an unfair advantage.}
\label{table:robustness_scores_resnet50_all}
\end{table}

\begin{table}[h]
\centering
\begin{adjustwidth}{-0.7cm}{}
    \begin{tabular}{cc|cccccccccc}
    \toprule
    Dataset & Method & Normal & FGSM$^1$ & FGSM$^2$ & JSMA & PGD$^1$ & PGD$^2$ & Deepfool & CW$_{L2}$ & CW$_{L\infty}$ & Square \\
    \hline
    \rowcolor{Gray}
    \multirow{8}{*}{CIFAR-10} & Plain & 94.96 & 58.08 & 44.12 & 77.92 & 9.04 & 0.00 & 3.60 & 1.32 & 24.24 & 56.32 \\
    \rowcolor{Gray}
    & Ensemble & 96.24 & 78.76 & 55.56 & 87.48 & 77.52 & 14.36 & 93.52 & 83.08 & 81.16 & 87.60 \\
    & TRADES   & 85.04 & 82.76 & 72.80 & 70.20 & 82.88 & 68.08 & 8.56 & 0.16 & 75.08 & 77.56 \\
    & VAT      & \textbf{94.36} & 80.24 & 64.40 & 83.84 & 76.76 & 10.96 & 3.60 & 2.12 & 36.28 & 82.68 \\
    & TTA      & 91.52 & 77.92 & 58.96 & 84.80 & 83.04 & 55.24 & 86.36 & 81.60 & 80.60 & 83.16 \\
    & ARF & 93.64 & 81.64 & 63.44 & 84.88 & 88.00 & 60.48 & \textbf{91.20} & 83.96 & 86.48 & 86.36 \\
    & TRADES + ARF & 82.12 & 81.00 & 75.24 & 77.24 & 81.16 & 74.04 & 66.00 & 67.12 & 78.88 & 79.00 \\
    & VAT + ARF & 93.40 & \textbf{88.92} & \textbf{76.52} & \textbf{90.60} & \textbf{89.76} & \textbf{76.36} & 88.88 & \textbf{85.48} & \textbf{87.84} & \textbf{89.00} \\
    \hline
    \rowcolor{Gray}
    \multirow{8}{*}{CIFAR-100} & Plain & 75.56 & 33.20 & 17.56 & 53.44 & 16.72 & 0.12 & 9.40 & 13.52 & 46.92 & 27.64 \\
    \rowcolor{Gray}
    & Ensemble & 79.56 & 60.48 & 29.84 & 59.52 & 69.64 & 42.60 & 78.12 & 54.96 & 72.00 & 63.72 \\
    & TRADES & 55.52 & 54.68 & 44.28 & 46.64 & 54.60 & 49.88 & 10.36 & 4.52 & 53.08 & 47.32 \\
    & VAT & \textbf{74.28} & 55.08 & 31.32 & 59.72 & 64.80 & 10.56 & 9.44 & 8.84 & 41.12 & 52.12 \\
    & TTA    & 68.48 & 51.16 & 32.60 & 59.04 & 62.12 & 48.16 & 65.16 & 48.96 & 62.16 & 56.48 \\
    & ARF & 70.60 & 55.40 & 35.92 & 61.20 & 66.16 & 53.16 & \textbf{67.64} & 53.80 & \textbf{66.48} & 59.52 \\
    & TRADES + ARF & 50.88 & 49.76 & 44.72 & 46.32 & 49.40 & 45.88 & 43.68 & 35.00 & 49.68 & 46.96 \\
    & VAT + ARF & 70.52 & \textbf{65.16} & \textbf{48.72} & \textbf{66.12} & \textbf{67.48} & \textbf{60.36} & 66.56 & \textbf{56.84} & 62.92 & \textbf{64.52} \\
    \hline
    \rowcolor{Gray}
    \multirow{8}{*}{SVHN} & Plain & 97.48 & 80.12 & 62.12 & 57.72 & 53.00 & 1.52 & 2.32 & 4.88 & 33.36 & 71.32 \\
    \rowcolor{Gray}
    & Ensemble & 98.08 & 90.04 & 72.04 & 87.80 & 92.72 & 64.80 & 84.12 & 83.12 & 88.12 & 93.24 \\
    & TRADES & 93.52 & 93.12 & 89.32 & 38.88 & 92.40 & 74.12 & 4.60 & 0.68 & 87.92 & 83.60 \\
    & VAT    & 94.64 & 89.60 & 86.24 & 75.72 & 71.20 & 12.64 & 3.32 & 12.64 & 56.32 & 78.72 \\
    & TTA    & 97.16 & 87.92 & 71.00 & 86.56 & 90.36 & 54.92 & 80.60 & 81.92 & 85.36 & 93.36 \\
    & ARF    & \textbf{97.24} & 88.80 & 71.72 & 87.60 & 91.08 & 61.56 & 82.68 & 82.32 & 86.28 & \textbf{93.68} \\ 
    & TRADES + ARF & 94.04 & 93.32 & \textbf{89.88} & 79.52 & 93.28 & \textbf{82.56} & 72.68 & 54.36 & 90.48 & 88.64 \\
    & VAT + ARF & 95.52 & \textbf{93.80} & 89.04 & \textbf{93.32} & \textbf{94.32} & 82.28 & \textbf{95.72} & \textbf{91.08} & \textbf{92.24} & 92.04 \\
    \hline
    \rowcolor{Gray}
    \multirow{8}{*}{\footnotesize Tiny ImageNet} & Plain & 66.56 & 25.88 & 11.76 & 34.48 & 30.44 & 0.24 & 9.08 & 14.56 & 40.60 & 32.00 \\
    \rowcolor{Gray}
    & Ensemble & 70.20 & 55.24 & 27.92 & 53.16 & 65.96 & 46.12 & 68.80 & 53.16 & 61.88 & 61.84 \\
    & TRADES   & 45.28 & 43.88 & 35.72 & 37.80 & 44.36 & 40.92 & 10.96 & 5.04 & 44.20 & 41.64 \\
    & VAT      & 64.40 & 43.96 & 23.84 & \textbf{53.24} & 48.92 & 4.68 & 10.16 & 26.80 & 51.84 & 46.44 \\
    & TTA      & 54.84 & 39.08 & 19.52 & 43.48 & 46.00 & 29.96 & 49.96 & 34.84 & 44.76 & 42.92 \\
    & ARF      & \textbf{57.12} & 41.80 & 22.16 & 45.80 & 50.72 & 35.04 & 51.28 & 37.76 & 47.68 & 46.52 \\
    & TRADES + ARF & 35.64 & 34.04 & 30.48 & 34.56 & 33.88 & 31.88 & 30.56 & 20.88 & 35.76 & 32.88 \\
    & VAT + ARF & 56.12 & \textbf{50.40} & \textbf{36.72} & 53.00 & \textbf{53.92} & \textbf{49.68} & \textbf{52.88} & \textbf{46.40} & \textbf{53.32} & \textbf{50.12} \\
    \bottomrule
    \end{tabular}
\end{adjustwidth}
 \caption{Comparison of accuracies (\%) for various classifiers on CIFAR-10, CIFAR-100, SVHN, and Tiny ImageNet trained on Resnet-101. All attacks are detailed in Section~4 in the main paper. We boldface the best results. Ensemble is presented just as a reference as it has an unfair advantage.}
 \label{table:robustness_scores_resnet101_all}
\end{table}

Overall, the results on these DNNs have the same trend as in Resnet-34. ARF's robustness is on par with TRADES and VAT, however when combined with VAT we surpass the robustness of the vanilla adversarial trained DNN by a large margin. In addition, contrary to TRADES, ARF exhibits very high normal accuracy.

\clearpage

\section{Test-time augmentations}
\label{sec:test_time_aug}
Here we detail all the transforms we used to generate our Test-Time Augmentations (TTAs) for our robust classification methods, as described in Section~3.1 in the main paper. We used different parameters for \textit{soft} transforms and \textit{hard} transforms in the ablation study in Section~5.2. Both sets of parameters are listed below. The main result in the paper, outside the aforementioned ablation study, were calculated only with the \textit{hard} set, which proved to achieve better performance in the ablation study. We denote the original and transformed images as $x$ and $x_t$, respectively. 

\begin{enumerate}
\item Rotation: Angle rotation of the image was randomized to be in $\textit{U}(-8^{\circ}, 8^{\circ})$ for \textit{soft} and $\textit{U}(-15^{\circ},15^{\circ})$ for \textit{hard}.
\item Translation: The image was allowed to shift horizontally and vertically up to 2 pixels in every direction for CIFAR-10, CIFAR-100, and SVHN, and up to 4 pixels for Tiny ImageNet. This transform behaves similarly for both \textit{soft} and \textit{hard}.
\item{Scale: We randomly selected a zoom in ($s > 1$) or a zoom out ($s < 1$).
The image was scaled with $s \sim \textit{U}(0.95, 1.05)$ for \textit{soft} and $s \sim \textit{U}(0.9, 1.1)$ for \textit{hard}.}
\item Mirror: The image was horizontally flipped with a probability of 0.5. This transform was omitted for SVHN dataset, and was the same for \textit{soft} and \textit{hard}.
\item{Brightness: Randomly increase/decrease brightness. Let $b$ denote the brightness factor; the transforms is defined as $x_t = b \cdot x$. We randomized $b \sim \textit{U}(0.8, 1.2)$ for \textit{soft} and $b \sim \textit{U}(0.6, 1.4)$ for \textit{hard}.}
\item Contrast: The contrast factor $c$ was distributed as $c \sim \textit{U}(0.85, 1.15)$ for \textit{soft} and as $c \sim \textit{U}(0.7, 1.3)$ for \textit{hard}. The transformed image after contrast is: $x_t = c \cdot x + (1-c) \cdot \mathbb{E}(x_G) \cdot \mathbbm{1}_{nxnx3}$, where $\mathbb{E}(x_G)$ is the mean pixel value on the gray-scale equivalent image and $\mathbbm{1}_{nxnx3}$ is a matrix as the size of the original image, filled with ones. The gray-scale image is defined as: $x_G = 0.2989 \cdot R + 0.587 \cdot G + 0.114 \cdot B$ where $(R,G,B)$ are the red, green, and blue channels of $x$, respectively.
\item Saturation: The saturation factor $sat$ was distributed as $sat \sim \textit{U}(0.75, 1.25)$ for \textit{soft} and as $sat \sim \textit{U}(0.5, 1.5)$ for \textit{hard}. It is defined as: $x_t = sat \cdot x + (1-sat) \cdot x_G$.
\item Hue: The hue factor $h$ was distributed as $h \sim \textit{U}(0.03, 0.03)$ for \textit{soft} and as $h \sim \textit{U}(0.06, 0.06)$ for \textit{hard}. The transform updates the hue in the Hue Saturation Value (HSV) representation by $h$.
\item Gamma: Applying gamma transform on the image. Each channel (r,g,b) on $x$ is transformed to $x_t[r,g,b] = x[r,g,b]^\gamma$, where $\gamma \sim \textit{U}(0.85, 1.15)$ for \textit{soft} and $\gamma \sim \textit{U}(0.7, 1.3)$ for \textit{hard}.
\item Blur: The blur transform convolutes the image with a 2D Gaussian kernel: $x_t = G_{2D}(u, v; \sigma_b) * x$, where $G_{2D}(u, v; \sigma_b) = \frac{1}{2\pi\sigma_b^2}\exp{\frac{-(u^2 +v^2)}{2\sigma_b^2}}$,
where $\sigma_b$ is uniformly distributed between $0.001$ and a positive constant value $\sigma_{bmax}$: $\sigma_b \sim \textit{U}(0.001, \sigma_{bmax})$.

We set $\sigma_{bmax}=0.25$ for \textit{soft} and  $\sigma_{bmax}=0.5$ for \textit{hard}.
\item Noise: The Noise transform adds a white Gaussian noise to the image, $x_t = x + n$, where n is sampled from $n \sim \textit{N}(0, \sigma)$. The standard deviation of the normal distribution is randomized in our algorithm to be $\sigma \sim \textit{U}(0, \sigma_{max})$. We set $\sigma_{max}=0.005$ in all our experiments (see Section~5.2 in the main paper).
\end{enumerate}

It is important to point out that for all the color transforms, geometric transforms (except Mirror) and Noise, the mean value of the transform change is zero, thus our generated TTAs are unbiased.

The transforms were carried out in the following order:
\begin{enumerate} 
    \item[A)] Applying all the color transforms ([5]-[9]). The order of the color transforms was randomized.
    \item[B)] Padding the image with the last value at the edge of the image. CIFAR-10, CIFAR-100, and SVHN were padded to 64x64x3 and Tiny ImageNet was padded to 128x128x3.
    \item[C)] Applying the random affine transform (transforms [1]-[3]).
    \item[D)] Blurring the image ([10]).
    \item[E)] Croppong the center of the image.
    \item[F)] Applying random horizontal flip (not for SVHN) ([4]).
    \item[G)] Adding noise ([11]).
\end{enumerate}


Some of our transforms were implemented using the TorchVision package of PyTorch (\cite{PYTORCH}).

\section{Hardware setup}
\label{sec:hardware}
We trained our Deep Neural Networks (DNNs), Resnet-34, Resnet-50, and Resnet-101, with a GPU of type NVIDIA GeForce RTX 2080 Ti. This GPU has 11 GB of VRAM. We used multi workers setup and utilized 4 threads of Intel Xeon Silver 4114 CPU. 

All the adversarial training with TRADES (\cite{TRADES}) required more memory, therefore these DNNs were trained on a different server using NVIDIA RTX A6000 GPU that has 48 GB of VRAM. For training with TRADES we used 4 threads of Intel Xeon Gold 5220R CPU.

All the attacks listed in Section~4 in the paper, including the adapted attacks, were carried out on a single NVIDIA GeForce RTX 2080 Ti GPU.
All the DNNs training, adversarial attacks, and evaluations were done using a single GPU.

The TTAs were generated on the CPU alone. After generated them, we fed them to the DNNs with a single forward pass (of 256 TTAs).

The random forest classifier was trained using 20 threads of Intel Xeon Silver 4114 CPU.

\section{Adversarial training}
\label{sec:adversarial_training}
\setcounter{table}{0}
\renewcommand{\thetable}{D\arabic{table}}
We trained some TRADES models for fewer train epochs since we observed this yields more robust classifiers. The normal and adversarial accuracies on CIFAR-10, CIFAR-100, SVHN, and Tiny ImageNet trained on Resnet-34 using TRADES , is shown in Table~\ref{table:trades_epochs}. The attacks listed in the table are PGD($L_\infty, \epsilon=0.01$), PGD($L_\infty, \epsilon=0.031$), and CW($L_\infty, \epsilon=0.031$), which are defined in Section~4 in the main paper, abbreviated to PGD$^1$, PGD$^2$, and CW$_{L_\infty}$, respectively.
Based on these results we trained all the adversarial robust TRADES DNN with 100, 100, 100, and 300 epochs for CIFAR-10, CIFAR-100, SVHN, and Tiny ImageNet, respectively.
The only exception was training Tiny ImageNet on Resnet-101 with TRADES which was very time consuming, therefore we trained it only for 100 epochs instead of 300 epochs.

\begin{table}[ht!]
\centering
\begin{tabular}{cc|cccc}
\toprule
Dataset & Epochs & Normal & PGD$^1$ & PGD$^2$ & CW$_{L_\infty}$ \\
\hline
\multirow{3}{*}{CIFAR-10} & 100 & 86.68 & \textbf{85.12} & \textbf{71.88} & \textbf{78.28} \\
& 200 & \textbf{87.08} & 84.00 & 67.96 & 74.36 \\
& 300 & 86.92 & 84.28 & 68.92 & 75.12 \\
\hline
\multirow{3}{*}{CIFAR-100} & 100 & \textbf{53.36} & \textbf{52.88} & 46.44 & \textbf{51.04} \\
& 200 & 53.00 & 52.00 & \textbf{47.28} & 50.20 \\
& 300 & 53.00 & 52.32 & 47.16 & 50.04 \\
\hline
\multirow{2}{*}{SVHN} & 100 & \textbf{92.48} & \textbf{90.28} & \textbf{70.88} & \textbf{82.36} \\
& 200 & 91.64 & 84.64 & 41.04 & 56.16 \\
\hline
\multirow{3}{*}{Tiny ImageNet} & 100 & 41.68 & 41.04 & 37.48 & 40.20 \\
& 200 & 43.88 & 43.08 & 37.96 & \textbf{42.60} \\
& 300 & \textbf{44.44} & \textbf{43.72} & \textbf{38.44} & 41.88 \\
\bottomrule
\end{tabular}
\caption{Normal and adversarial accuracies (\%) for adversarially robust DNNs trained with TRADES on Resnet-34, for various number of epochs.}
\label{table:trades_epochs}
\end{table}

We trained the VAT models with the same number of epochs as the vanilla Resnets: 300, 300, 200, and 300 epochs for CIFAR-10, CIFAR-100, SVHN, and Tiny ImageNet, respectively. Unlike TRADES, the VAT robustness did not degrade in the late epochs, as shown in Table~\ref{table:vat_epochs}.

\begin{table}[ht!]
\centering
\begin{tabular}{cc|cccc}
\toprule
Dataset & Epochs & Normal & PGD$^1$ & PGD$^2$ & CW$_{L_\infty}$ \\
\hline
\multirow{3}{*}{CIFAR-10} & 100 & 93.04 & 81.12 & 15.00 & 34.32  \\
& 200 & \textbf{94.00} & 81.60 & 19.44 & 49.68 \\
& 300 & \textbf{94.00} & \textbf{82.12} & \textbf{20.08} & \textbf{49.80} \\
\hline
\multirow{3}{*}{CIFAR-100} & 100 & 66.88 & 58.48 & 12.12 & 34.24 \\
& 200 & 70.84 & 62.12 & 14.96 & 42.44 \\
& 300 & \textbf{70.92} & \textbf{63.00} & \textbf{15.20} & \textbf{44.36} \\
\hline
\multirow{2}{*}{SVHN} & 100 & 81.92 & 63.77 & 19.23 & 45.36 \\
& 200 & \textbf{94.90} & \textbf{85.00} & \textbf{42.35} & \textbf{64.68} \\
\hline
\multirow{3}{*}{Tiny ImageNet} & 100 & 50.69 & 49.03 & 29.90 & 41.84 \\
& 200 & 54.06 & 51.86 & 32.28 & 45.38 \\
& 300 & \textbf{54.67} & \textbf{52.20} & \textbf{32.48} & \textbf{45.94} \\
\bottomrule
\end{tabular}
\caption{Normal and adversarial accuracies (\%) for adversarially robust DNNs trained with VAT on Resnet-34, for various number of epochs.}
\label{table:vat_epochs}
\end{table}

To optimize the VAT training, we set $\alpha=1$ as suggested in \cite{Virtual_Adversarial_Training}, and experimented with different values of $\epsilon$, as listed in Table~\ref{table:vat_epsilons}. Based on these results, we chose to optimize VAT for max robustness on PGD$^2$ and thus selected $\epsilon$=1, 1, 3, 1 for CIFAR-10, CIFAR-100, SVHN, and Tiny ImageNet, respectively.

\begin{table}[h]
\centering
\begin{tabular}{cc|cccc}
\toprule
Dataset & $\epsilon$ & Normal & PGD$^1$ & PGD$^2$ & CW$_{L_\infty}$ \\
\hline
\multirow{4}{*}{CIFAR-10} & 0.5 & \textbf{95.16}	& 66.44	&	1.92	&	27.52 \\
& 1 & 94.00& 	\textbf{82.12}	&	\textbf{20.08}	&	\textbf{49.80} \\
& 2 & 94.76	& 26.28	& 	0.16	& 	22.72 \\
& 8 &  89.80&	72.08	&	11.28&		34.28\\
\hline
\multirow{4}{*}{CIFAR-100} & 0.5 & \textbf{74.48} &	54.80	&	3.32	&	43.08\\
& 1 & 70.92	& \textbf{63.00}	&	\textbf{15.20}	&	\textbf{44.36} \\
& 2 & 70.48	& 36.68	&	1.48	&	35.64 \\
& 8 & 62.28	& 55.68	&	15.16	&	31.20 \\
\hline
\multirow{3}{*}{SVHN} & 0.5 & \textbf{95.65}	& 82.70	&	18.35	&	44.87 \\
& 1 & 95.19	& 53.17	&	5.95	&	\textbf{77.18} \\
& 3 & 94.90	& \textbf{85.00}	&	\textbf{42.35}	&	64.68 \\
\hline
\multirow{4}{*}{Tiny ImageNet} & 0.5 & 57.82&	53.51&		17.71	&	42.32\\
& 1 & 54.67	& \textbf{52.20}	&	\textbf{32.48} &		\textbf{45.94} \\
& 2 & \textbf{58.11}	& 38.50	&	1.40&		42.39 \\
& 8 & 53.67	& 47.58	& 	7.30&		43.01 \\
\bottomrule
\end{tabular}
\caption{Normal and adversarial accuracies (\%) for adversarially robust DNNs trained with VAT on Resnet-34, for various number of $\epsilon$ values, as defined in \cite{Virtual_Adversarial_Training}.}
\label{table:vat_epsilons}
\end{table}

\clearpage

\section{Alternative classifiers}
\label{sec:alternative_cls}
\setcounter{table}{0}
\renewcommand{\thetable}{E\arabic{table}}
We tested three different simple models instead of our random forest classifier in ARF: Logistic regression, linear SVM, and SVM with an RBF kernel. Since our datasets are multi class, we set the classification strategy to be one-vs-rest. Table~\ref{table:lr_and_svm} shows results of normal and adversarial accuracies on CIFAR-10, trained on Resnet-34, and attacked by all the non adaptive attacks described in Section~4 in the main paper.

We observe that the random forest classifier achieves much better performance than the linear classifiers, and overall it is slightly better than SVM with RBF kernel. In addition, the ARF achieves the highest normal accuracy among all the classifiers we tested. Since SVM with RBF has approximately the same computation run time as random forest, there is no reason to favor it over random forest.

\begin{table}[h]
\centering
\resizebox{\columnwidth}{!}{
\begin{tabular}{c|ccccccccccc}
\toprule
Classifier &  Normal & FGSM$^1$ & FGSM$^2$ & JSMA & PGD$^1$ & PGD$^2$ & Deepfool & CW$_{L_2}$ & CW$_{L_\infty}$ & Square & Boundary \\
\hline
Logistic regression  & 92.68 & 82.40 & 68.72 & 82.44 & 86.24 & 72.32 & 84.68 & 81.20 & 80.36 & 83.88 & 89.60 \\
SVM (linear)         & 89.40 & 79.76 & 67.48 & 79.84 & 81.31 & 65.16 & 79.12 & 76.92 & 75.52 & 80.24 & 87.60 \\
SVM (RBF)            & 93.52 & 83.52 & 69.76 & 84.68 & 89.76 & \textbf{78.16} & \textbf{87.36} & \textbf{84.80} & 85.28 & 87.08 & \textbf{91.60} \\
Random forest        & \textbf{93.76} & \textbf{83.72} & \textbf{70.20} & \textbf{85.28} & \textbf{90.32} & 77.88 & \textbf{87.36} & 84.36 & \textbf{85.64} & \textbf{87.84} & 91.20 \\
\bottomrule
\end{tabular}
}
\vspace{-0.11in}
\caption{Normal and adversarial accuracies (\%) on CIFAR-10 when training logistic regression or SVM compared to our proposed random forest classifier.}
\label{table:lr_and_svm}
\end{table}

\section{Transferability}
\label{sec:transferability}
\setcounter{table}{0}
\renewcommand{\thetable}{F\arabic{table}}
We show that our ARF defense is characterized with excellent transferability, being able to generalize to new (unseen) attacks. Table~\ref{table:global_trans} compares between two different setups of fitting and testing ARF. The top row shows the accuracies we presented in Table~1 in the main paper, when training ARF on all the attacks (FGSM$^1$, FGSM$^2$, JSMA, PGD$^1$, PGD$^2$, Deepfool, CW$_{L_2}$, CW$_{L_\infty}$, Square, and Boundary) with a global random forest model, obtained by fitting it on all the aforementioned attacks. The second row shows ARF accuracy using the Leave-One-Out Cross-Validation (LOOCV) procedure, where we fit the random forest on all the attacks except the attack we wish to test it on. For example, we calculate the adversarial accuracies on images generated by FGSM$^1$ and FGSM$^2$ after fitting the random forest on images generated by JSMA, PGD$^1$, PGD$^2$, Deepfool, CW$_{L_2}$, CW$_{L_\infty}$, Square, and Boundary. Table~\ref{table:LOOCV} lists explicitly which attacks were used to fit the random forest for each tested attack. Since this cross-validation method fits seven random forest models, the displayed normal accuracy is their calculated mean and standard deviation.

\begin{table}[h]
\centering
\begin{adjustwidth}{-1.23cm}{}
    \begin{tabular}{c|ccccccccccc}
    \toprule
    Random forest fitting & Normal & FGSM$^1$ & FGSM$^2$ & JSMA & PGD$^1$ & PGD$^2$ & Deepfool & CW$_{L_2}$ & CW$_{L_\infty}$ & Square & Boundary \\
    \hline
    Global & 93.76 & 83.72 & 70.20 & 85.28 & 90.32 & 77.88 & 87.36 & 84.36 & 85.64 & 87.84 & 91.20 \\
    LOOCV & 93.71 $\pm$ 0.07 & 83.48 & 69.48 & 84.92 & 90.16 & 78.16 & 87.32 & 84.92 & 85.24 & 87.92 & 91.60 \\
    \bottomrule
    \end{tabular}
\end{adjustwidth}
\vspace{-0.11in}
\caption{Normal and adversarial accuracies (\%) on CIFAR-10 using different setups for fitting the random forest. The top row is the method shown in the main paper, where the random forest is fitted and tested on all the non adaptive attacks. The bottom row shows results for the Leave-One-Out Cross-Validation (LOOCV) method, where the tested attack is excluded from the random forest fitting.}
\label{table:global_trans}
\end{table}

\begin{table}[h]
\centering
    \begin{tabular}{c|cccccccccc}
    
    Tested attack & FGSM$^1$ & FGSM$^2$ & JSMA & PGD$^1$ & PGD$^2$ & Deepfool & CW$_{L_2}$ & CW$_{L_\infty}$ & Square & Boundary\\
    \hline
    FGSM$^1$ & & & \checkmark & \checkmark & \checkmark & \checkmark & \checkmark & \checkmark & \checkmark & \checkmark \\
    \hline
    FGSM$^2$ &  & & \checkmark & \checkmark & \checkmark & \checkmark & \checkmark & \checkmark & \checkmark & \checkmark \\
    \hline
    JSMA & \checkmark & \checkmark & & \checkmark & \checkmark & \checkmark & \checkmark & \checkmark & \checkmark & \checkmark\\
    \hline
    PGD$^1$ & \checkmark & \checkmark & \checkmark & &   & \checkmark & \checkmark & \checkmark & \checkmark & \checkmark\\
    \hline
    PGD$^2$ & \checkmark & \checkmark & \checkmark &  & & \checkmark & \checkmark & \checkmark & \checkmark & \checkmark\\
    \hline
    Deepfool & \checkmark & \checkmark & \checkmark & \checkmark & \checkmark & & \checkmark & \checkmark & \checkmark & \checkmark\\
    \hline
    CW$_{L_2}$ & \checkmark & \checkmark & \checkmark & \checkmark & \checkmark & \checkmark & & & \checkmark & \checkmark\\
    \hline
    CW$_{L_\infty}$ & \checkmark & \checkmark & \checkmark & \checkmark & \checkmark & \checkmark & & & \checkmark & \checkmark\\
    \hline
    Square & \checkmark & \checkmark & \checkmark & \checkmark & \checkmark & \checkmark & \checkmark & \checkmark  & & \checkmark\\
    \hline
    Boundary & \checkmark & \checkmark & \checkmark & \checkmark & \checkmark & \checkmark & \checkmark & \checkmark  & \checkmark &\\
    \bottomrule
    \end{tabular}
\vspace{-0.11in}
\caption{Each row displays which attacks were employed to fit the random forest on the tested attack using the LOOCV procedure.}
\label{table:LOOCV}
\end{table}

\clearpage
\section{BPDA attack}
\label{sec:bpda}
In the white-box settings, we utilized the Backward Pass Differentiable Approximation (BPDA) attack by \cite{Athalye2018ObfuscatedGG}. More specifically, we employed the generalized BPDA and replaced the non-differential random forest model with a substitute model that can derive gradients. To that end we used knowledge distillation (\cite{Hinton2015DistillingTK}) to train a substitute MLP to mimic the random forest functionality. We used a six layer MLP with the following dimensions:
\begin{itemize}
  \item For CIFAR-10 and SVHN: $N \cdot C$ $\rightarrow$ $N \cdot C$ $\rightarrow$ $\frac{N \cdot C}{2}$ $\rightarrow$ $\frac{N \cdot C}{4}$ $\rightarrow$ $\frac{N \cdot C}{8}$ $\rightarrow$ $\frac{N \cdot C}{16}$ $\rightarrow$ $C$.
  \item For CIFAR-100 and Tiny ImageNet: $N \cdot C$ $\rightarrow$ $\frac{N \cdot C}{10}$ $\rightarrow$ $\frac{N \cdot C}{20}$ $\rightarrow$ $\frac{N \cdot C}{40}$ $\rightarrow$ $\frac{N \cdot C}{80}$ $\rightarrow$ $\frac{N \cdot C}{160}$ $\rightarrow$ $C$.
\end{itemize}
$N$ and $C$ are the TTA size and the number of classes in the dataset, respectively. Since CIFAR-100 and Tiny-ImageNet have higher number of logits, we shrink their layer size in the MLP faster to limit the number of parameters.

After each linear layer we used a batch normalization layer and Relu activation (in this order), except for the last layer. We trained this MLP with TTA logits obtained solely from the \textit{test} set (used for fitting the random forest), and kept the \textit{test-val} hidden. We train this MLP using the KL divergence loss; we did not add the cross entropy loss (with the ground truth label) to the training since our goal is to mimic the random forest gradients with the highest fidelity, and not to improve classification accuracy.

After the MLP was trained, we used the BPDA attack on the hybrid model encapsulating the original DNN and the substituted MLP, connected in tandem. All our robustness results, including on the BPDA attack, show accuracy calculated for adversarial imaged generated from the \textit{test-val} set. It should be emphasized that the above hybrid model (DNN+MLP) was only used to generate the adversarial images. For evaluating robustness we pass these images to the ARF model (DNN with random forest).

The EoT attack (\cite{Athalye2018SynthesizingRA}) was not used in our experiments because this attack requires a white-box threat model with differential loss. Thus, we could not differentiate the expected value of a loss at the output of the random forest, over our TTA transform distribution. In any rate, we showed in the paper that the transformations alone do not provide protection against adaptive white-box attacks, since our A-PGD attack greatly attenuates the ARF robustness (see Section~5.3 in the paper).

\section{TTA size ablation}
\label{sec:tta_size_ablation}
\setcounter{figure}{0}
\renewcommand{\thefigure}{H\arabic{figure}}

\begin{figure}[h]
\centering
\includegraphics[width=\linewidth]{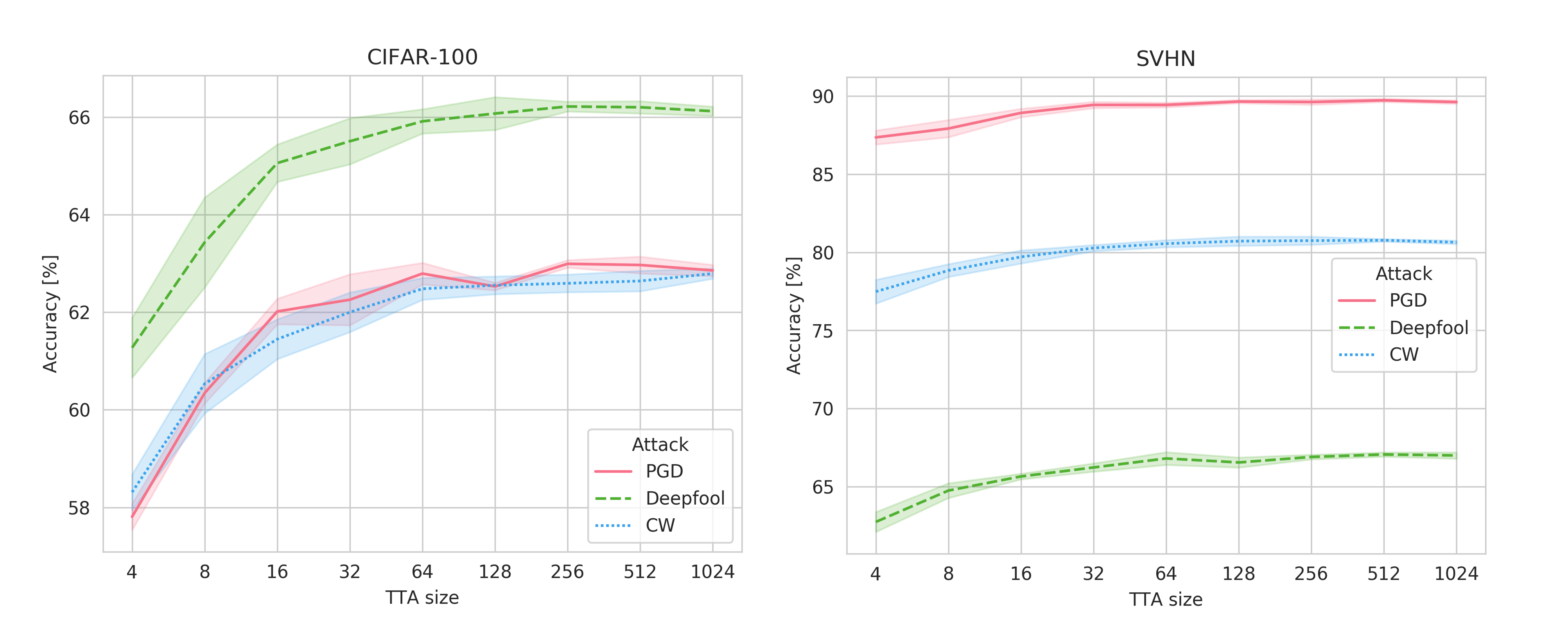}
\caption{Ablation study on the number of generated TTAs ($N$). We calculate the adversarial accuracies on CIFAR-100 and SVHN for three attacks as a function of $N$ (logarithmic scale).}
\label{fig:tta_size_ablation_cifar100_svhn}
\end{figure}

Here we repeat the TTA size ablation test in Section~5.2 for CIFAR-100 and SVHN datasets. We plot the adversarial accuracies for PGD$^1$, Deepfool, and CW$_{L_2}$ in a logarithmic scale (Figure~\ref{fig:tta_size_ablation_cifar100_svhn}), and show that $N=256$ TTAs are sufficient also for these datasets.

\clearpage
\section{$L_2$ distortion}
\label{sec:L2_distortion}
\setcounter{table}{0}
\setcounter{figure}{0}
\renewcommand{\thetable}{I\arabic{table}}
\renewcommand{\thefigure}{I\arabic{figure}}

Table~\ref{table:L2_distortions} reports the mean $L_2$ distortion of the adversarial images, generated by Deepfool (gray-box), CW$_{L_2}$ (gray-box), A-PGD (adaptive gray-box), Boundary (black-box), and BPDA (white-box) attacks. The $L_2$ distortion of an adversarial image $x'$ from the natural image $x$ is defined by $d_{L_2}(x, x') = ||x - x'||_2$. In our average distortion calculation we consider only adversarial images that fooled the defense, meaning, the DNN classified $x$ correctly but the defense method misclassified $x'$.

This measure is interesting because high $L_2$ distortion value correlates to perceptible noises on the images, thus invalidates the attack since humans can easily notice it. The above is relevant especially for the unbounded attacks, Deepfool, CW$_{L_2}$, and Boundary (in our experiments), which are not constrained by the $L_2$ norm.

We observe that in the majority of cases, the highest mean $L_2$ distortion was obtained by using TRADES, either alone or combined with ARF. This finding supports our conclusions from the main paper advocating the use of ARF on top of an adversarially trained DNN.

\begin{table}[ht]
\centering
\begin{tabular}{cc|cccccc}
\toprule
\multirow{2}{*}{Dataset} & \multirow{2}{*}{Attack} & \multirow{2}{*}{Plain} & \multirow{2}{*}{TRADES} & \multirow{2}{*}{VAT} & \multirow{2}{*}{ARF} & TRADES+ & VAT+ \\
& & & & & & ARF & ARF \\
\hline
\multirow{5}{*}{CIFAR-10}       & Deepfool    & 0.58 & \textbf{1.48} & 1.46 & 0.30 & 1.08 & 0.56 \\
&                                 CW$_{L_2}$  & 0.80 & 1.99 & 1.67 & 2.31 & 1.98 & \textbf{2.79} \\
&                                 Boundary    & 0.25 & \textbf{1.47} & 0.89 & 0.11 & 1.37 & 0.68 \\
&                                 A-PGD       & 1.36 & 1.60 & 1.52 & 1.35 & \textbf{1.61} & 1.52 \\
&                                 BPDA        & 1.27 & 1.36 & 1.36 & 1.27 & \textbf{1.38} & 1.36 \\
\hline
\multirow{5}{*}{CIFAR-100}      & Deepfool    & 0.20 & \textbf{0.85} & 0.54 & 0.07 & 0.50 & 0.20 \\
&                                 CW$_{L_2}$  & 1.67 & 2.92 & 1.89 & 2.51 & \textbf{3.01} & 2.31 \\
&                                 Boundary    & 0.41 & \textbf{2.29} & 1.07 & 0.25 & 1.99 & 0.77 \\
&                                 A-PGD       & 1.40 & \textbf{1.61} & 1.54 & 1.40 & \textbf{1.61} & 1.54 \\
&                                 BPDA        & 1.24 & \textbf{1.41} & 1.36 & 1.24 & 1.39 & 1.34 \\
\hline
\multirow{5}{*}{SVHN}           & Deepfool    & \textbf{1.29} & 1.07 & 0.94 & 1.18 & 1.09 & 0.51 \\
&                                 CW$_{L_2}$  & 1.10 & 1.37 & 1.52 & \textbf{2.50} & 1.32 & 2.01 \\
&                                 Boundary    & 0.43 & 0.97 & \textbf{1.18} & 0.16 & 0.61 & 0.63 \\
&                                 A-PGD       & 1.35 & \textbf{1.57} & 1.46 & 1.34 & 1.54 & 1.47 \\
&                                 BPDA        & 1.22 & \textbf{1.36} & 1.31 & 1.21 & \textbf{1.36} & 1.32 \\
\hline
\multirow{5}{*}{Tiny ImageNet}  & Deepfool    & 0.40 & \textbf{1.39} & 1.22 & 0.20 & 0.81 & 0.73 \\
&                                 CW$_{L_2}$  & 2.19 & \textbf{5.22} & 4.80 & 2.39 & 4.23 & 4.75 \\
&                                 Boundary    & 1.49 & \textbf{7.22} & 5.34 & 1.03 & 2.65 & 3.07 \\
&                                 A-PGD       & 2.86 & 3.19 & 3.18 & 2.87 & \textbf{3.20} & 3.17 \\
&                                 BPDA        & 2.44 & 2.80 & \textbf{2.86} & 2.44 & 2.82 & 2.83 \\
\bottomrule
\end{tabular}
\caption{Mean $L_2$ distortion values for adversarial images generated by selected attacks, on a vanilla Resnet34 (Plain), adversarially trained Resnet34 (TRADES/VAT), our ARF defense, and a combination of TRADES/VAT with our ARF.}
\label{table:L2_distortions}
\end{table}

\clearpage
\section{Visual perceptibility}
\label{sec:visual}
\setcounter{table}{0}
\setcounter{figure}{0}
\renewcommand{\thetable}{J\arabic{table}}
\renewcommand{\thefigure}{J\arabic{figure}}
In Section 5.3 in the main paper we show that our ARF defense is susceptible to the BPDA attack, an adaptive white-box attack that was customly tailored to circumvent our specific random forest classifier. In this section we show that this attack fails to generate imperceptible images. We display some images generated using BPDA against ARF and show that a human observer can easily detect an unusual distortion in them.

\begin{figure}[h]
\centering
\includegraphics[width=0.8\linewidth]{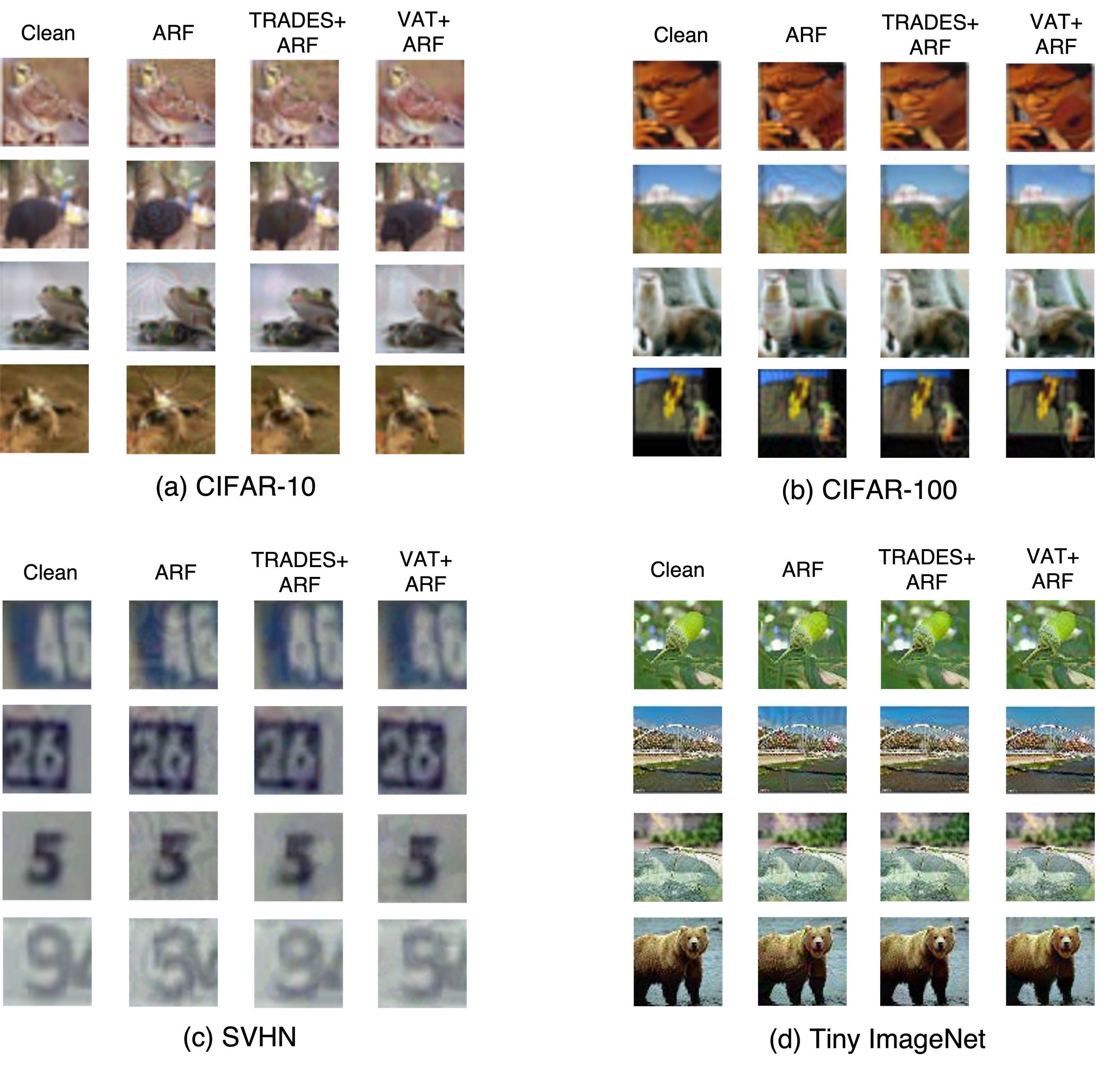}
\caption{Adversarial images generated by BPDA circumventing our ARF defense. TRADES+ARF and VAT+ARF correspond to our ARF defense when applied on top of an adversarially trained DNN, TRADES/VAT, respectively. Adversarial images that fool our ARF defense can be easily spotted by the naked eye.}
\label{fig:visualizations}
\end{figure}

Figure~\ref{fig:visualizations} exhibits clean images and adversarial images generated by BPDA for CIFAR-10, CIFAR-100, SVHN, and Tiny ImageNet. "Clean" column corresponds to natural (undistorted) images; "ARF" column denotes images that fool our ARF defense; "TRADES+ARF" and "VAT+ARF" columns display images that fool our ARF defense when combined with TRADES and VAT adversarially trained DNNs, respectively. For a fair comparison, we show only images that successfully fool all the three defenses, meaning, the DNN classified the clean image successfully but the adversarial image was able to flip the label despite our random forest classifier.

We note that the most visible noises correspond to attacks on the vanilla ARF method, without incorporating TRADES/VAT into it. This observation is counterintuitive to the reported accuracies in Section~5.3 in the main paper that show better robustness of ARF when combined with TRADES/VAT. In addition, Section~\ref{sec:L2_distortion} exhibits lower $L_2$ distortion for the vanilla ARF defense on the BPDA compared to TRADES+ARF and VAT+ARF. Nonetheless, these visible distortions decrease the efficacy of BPDA towards our defense.


\end{document}